\documentclass[11pt,a4paper]{article}
\usepackage[hyperref]{emnlp2020}
\usepackage{times}
\usepackage{latexsym}

\usepackage[utf8x]{inputenc}
\usepackage[thaicjk,english]{babel}
\addto\extrasthaicjk{\fontencoding{C90}\selectfont}
\makeatletter
\@namedef{opt@inputenc.sty}{utf8}
\makeatother
\usepackage{CJKutf8}
\usepackage{amsmath}
\usepackage{amsthm}
\usepackage{amsfonts}
\usepackage{amssymb}
\usepackage{mathtools}

\usepackage{graphicx}
\usepackage{booktabs}
\usepackage{tabularx}
\usepackage{tabulary}
\usepackage{color, colortbl}
\usepackage{xspace}
\usepackage{xcolor}
\usepackage{adjustbox}
\usepackage{subcaption}
\usepackage{multirow}
\usepackage{comment}
\usepackage{microtype}
\usepackage{arydshln}
\usepackage{tikz}

\definecolor{Gray}{gray}{0.92}
\newcolumntype{Y}{>{\centering\arraybackslash}X}

\usepackage{url}
\usepackage{enumitem}
\fboxsep=-\fboxrule
\usepackage{scalerel}

\newcommand{\en}{{\textsc{en}}\xspace}
\newcommand{\de}{{\textsc{de}}\xspace}
\newcommand{\fin}{{\textsc{fi}}\xspace}
\newcommand{\tr}{{\textsc{tr}}\xspace}
\newcommand{\ru}{{\textsc{ru}}\xspace}
\newcommand{\zh}{{\textsc{zh}}\xspace}

\newcommand{\ita}{{\textsc{it}}\xspace}

\newcommand{\norm}[1]{\left\lVert#1\right\rVert}

\newcommand{\mbert}{{\textsc{multi}}\xspace}
\newcommand{\monobert}{{\textsc{mono}}\xspace}
\newcommand{\nospec}{{\textsc{nospec}}\xspace}
\newcommand{\all}{{\textsc{all}}\xspace}
\newcommand{\cls}{{\textsc{withcls}}\xspace}
\newcommand{\tens}{{\textsc{10}}\xspace}
\newcommand{\huns}{{\textsc{100}}\xspace}
\newcommand{\aoc}{{\textsc{aoc}}\xspace}
\newcommand{\iso}{{\textsc{iso}}\xspace}

\aclfinalcopy 

\newcommand{\ignore}[1]{}

\usepackage{comments}
\newauthor[IV]{Ivan}{yellow}

\title{Probing Pretrained Language Models for Lexical Semantics}

\author{
Ivan Vuli\'{c}$^{\spadesuit}$~ Edoardo M. Ponti$^{\spadesuit}$~ Robert Litschko$^\diamondsuit$~ Goran Glava\v{s}$^\diamondsuit$~ Anna Korhonen$^{\spadesuit}$
 \\
$^\spadesuit$Language Technology Lab, University of Cambridge, UK\\
$^\diamondsuit$Data and Web Science Group, University of Mannheim, Germany\\ 
{ \small \tt \{iv250,ep490,alk23\}@cam.ac.uk} \\
{\small \tt \{goran,litschko\}@informatik.uni-mannheim.de
}
}

\date{}

\begin{document}
\maketitle

\begin{abstract}
The success of large pretrained language models (LMs) such as BERT and RoBERTa has sparked interest in probing their representations, in order to unveil what types of knowledge they implicitly capture. While prior research focused on morphosyntactic, semantic, and world knowledge, it remains unclear to which extent LMs also derive \textit{lexical type-level knowledge} from words in context. In this work, we present a \textit{systematic empirical analysis} across six typologically diverse languages and five different lexical tasks, addressing the following questions: \textbf{1)} How do different lexical knowledge extraction strategies (monolingual versus multilingual source LM, out-of-context versus in-context encoding, inclusion of special tokens, and layer-wise averaging) impact performance? How consistent are the observed effects across tasks and languages? \textbf{2)} Is lexical knowledge stored in few parameters, or is it scattered throughout the network? \textbf{3)} How do these representations fare against traditional static word vectors in lexical tasks? \textbf{4)} Does the lexical information emerging from independently trained monolingual LMs display latent similarities? Our main results indicate patterns and best practices that hold universally, but also point to prominent variations across languages and tasks. Moreover, we validate the claim that lower Transformer layers carry more type-level lexical knowledge, but also show that this knowledge is distributed across multiple layers.
\end{abstract}


\section{Introduction and Motivation}
\label{s:introduction}
Language models (LMs) based on deep Transformer networks \cite{Vaswani:2017nips}, pretrained on unprecedentedly large amounts of text, offer unmatched performance in virtually every NLP task \cite{Qiu:2020arxiv}. Models such as BERT \cite{devlin2018bert}, RoBERTa \cite{Liu:2019roberta}, and T5 \cite{Raffel:2019:arxiv} replaced task-specific neural architectures that relied on \textit{static} word embeddings \cite[WEs;][]{Mikolov:2013nips,Pennington:2014emnlp,Bojanowski:2017tacl}, where each word is assigned a single (type-level) vector.


While there is a clear consensus on the effectiveness of pretrained LMs, a body of recent research has aspired to understand \textit{why} they work \cite{Rogers:2020arxiv}. State-of-the-art models are ``probed'' to shed light on whether they capture task-agnostic linguistic knowledge and structures \cite{Liu:2019naacl,Belinkov:2019tacl,Tenney:2019acl}; e.g., they have been extensively probed for syntactic knowledge \cite[\textit{inter alia}]{Hewitt:2019naacl,Jawahar:2019acl,Kulmizev:2020acl,Chi:2020acl} and morphology \cite{Edmiston:2020arxiv,Hofmann:2020arxiv}.

In this work, we put focus on uncovering and understanding \textit{how and where lexical semantic knowledge} is coded in state-of-the-art LMs. While preliminary findings from \citet{Ethayarajh:2019emnlp} and \citet{Vulic:2020msimlex} suggest that there is a wealth of lexical knowledge available within the parameters of BERT and other LMs, \textit{a systematic empirical study across different languages} is currently lacking. 

We present such a study, spanning six typologically diverse languages for which comparable pretrained BERT models \textit{and} evaluation data are readily available. We dissect the pipeline for extracting lexical representations, and divide it into crucial components, including: the underlying source LM, the selection of subword tokens, external corpora, and which Transformer layers to average over. Different choices give rise to different \textit{extraction configurations} (see Table~\ref{tab:config}) which, as we empirically verify, lead to large variations in task performance.

We run experiments and analyses on five diverse lexical tasks using standard evaluation benchmarks: lexical semantic similarity (LSIM), word analogy resolution (WA), bilingual lexicon induction (BLI), cross-lingual information retrieval (CLIR), and lexical relation prediction (RELP). The main idea is to aggregate lexical information into static type-level ``BERT-based'' word embeddings and plug them into ``the classical NLP pipeline'' \cite{Tenney:2019acl}, similar to traditional static word vectors. The chosen tasks can be seen as ``lexico-semantic probes'' providing an opportunity to simultaneously \textbf{1)} evaluate the richness of lexical information extracted from different parameters of the underlying pretrained LM on intrinsic (e.g., LSIM, WA) and extrinsic lexical tasks (e.g., RELP); \textbf{2)} compare different type-level representation extraction strategies; and \textbf{3)} benchmark ``BERT-based'' static vectors against traditional static word embeddings such as fastText \cite{Bojanowski:2017tacl}.

Our study aims at providing answers to the following key questions: \textbf{Q1)} Do lexical extraction strategies generalise across different languages and tasks, or do they rather require language- and task-specific adjustments?; \textbf{Q2)} Is lexical information concentrated in a small number of parameters and layers, or scattered throughout the encoder?; \textbf{Q3)} Are ``BERT-based'' static word embeddings competitive with traditional word embeddings such as fastText?; \textbf{Q4)} Do monolingual LMs independently trained in multiple languages learn structurally similar representations for words denoting similar concepts (i.e., translation pairs)?


We observe that different languages and tasks indeed require distinct configurations to reach peak performance, which calls for a careful tuning of configuration components according to the specific task--language combination at hand (Q1). However, several universal patterns emerge across languages and tasks. For instance, lexical information is predominantly concentrated in lower Transformer layers, hence excluding higher layers from the extraction achieves superior scores (Q1 and Q2). Further, representations extracted from single layers do not match in accuracy those extracted by averaging over several layers (Q2). While static word representations obtained from monolingual LMs are competitive or even outperform static fastText embeddings in tasks such as LSIM, WA, and RELP, lexical representations from massively multilingual models such as multilingual BERT (mBERT) are substantially worse (Q1 and Q3). We also demonstrate that translation pairs indeed obtain similar representations (Q4), but the similarity depends on the extraction configuration, as well as on the typological distance between the two languages.


\section{Lexical Representations from Pretrained Language Models}
\label{s:methodology}
Classical static word embeddings \citep{bengio2003neural,Mikolov:2013nips,Pennington:2014emnlp} are grounded in distributional semantics, as they infer the meaning of each word type from its co-occurrence patterns. However, LM-pretrained Transformer encoders have introduced at least two levels of misalignment with the classical approach \cite{peters2018deep,devlin2018bert}. First, representations are assigned to word \textit{tokens} and are affected by the current context and position within a sentence \citep{Mickus:2020arxiv}. Second, tokens may correspond to subword strings rather than complete word forms. This begs the question: do pretrained encoders still retain a notion of lexical concepts, abstracted from their instances in texts?

Analyses of lexical semantic information in large pretrained LMs have been limited so far, focusing only on the English language and on the task of word sense disambiguation. \citet{reif2019visualizing} showed that senses are encoded with finer-grained precision in \textit{higher} layers, to the extent that their representation of the same token tends not to be self-similar across different contexts \citep{Ethayarajh:2019emnlp,Mickus:2020arxiv}. As a consequence, we hypothesise that abstract, type-level information could be codified in \textit{lower} layers instead. However, given the absence of a direct equivalent to a static word type embedding, we still need to establish \textit{how to extract} such type-level information.

In prior work, contextualised representations (and attention weights) have been interpreted in the light of linguistic knowledge mostly through \textit{probes}. These consist in learned classifier predicting annotations like POS tags \citep{pimentel2020informationtheoretic} and word senses \citep{peters2018deep,reif2019visualizing,chang-chen-2019-word}, or linear transformations to a space where distances mirror dependency tree structures \citep{Hewitt:2019naacl}.\footnote{The interplay between the complexity of a probe and its accuracy, as well as its effect on the overall procedure, remain controversial \citep{pimentel2020informationtheoretic,voita2020informationtheoretic}.}

\begin{figure}[!t]
    \centering
    \includegraphics[width=1.0\columnwidth]{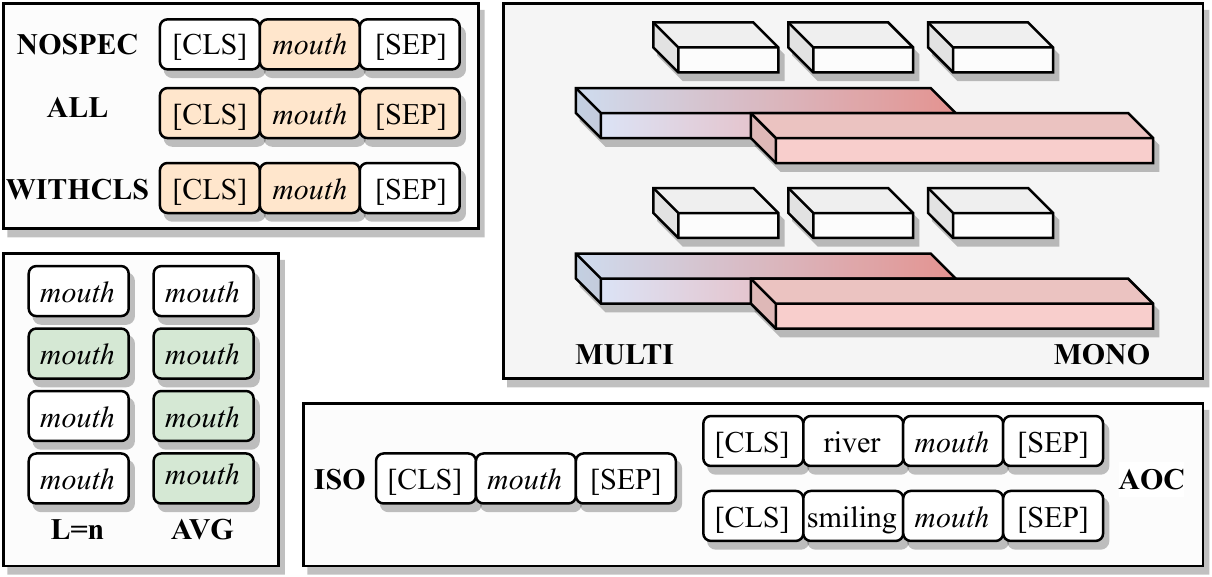}
    \caption{Illustration of the components denoting adopted extraction strategies, including source LM (top right), presence of context (bottom right), special tokens (top left), and layer-wise averaging (bottom left).}
    \vspace{-0mm}
    \label{fig:configs}
\end{figure}

\begin{table*}[t]
	\centering
    \def\arraystretch{1.1}
    {\footnotesize
	\begin{tabularx}{\textwidth}{l X l}
	    \toprule
		{\bf Component} & \textit{Label} & \textit{Short Description} \\
	    \cmidrule(lr){2-2} \cmidrule(lr){3-3}
		\multirow{2}{*}{\bf Source LM} & {\monobert} & {Language-specific (i.e., monolingually pretrained) BERT}\\
		& {\mbert} & {Multilingual BERT, pretrained on 104 languages (with shared subword vocabulary)}\\
        \cmidrule(lr){2-3}
		
		\multirow{2}{*}{\bf Context} & {\iso} & {Each vocabulary word $w$ is encoded \textit{in isolation}, without any external context} \\
		& {\aoc-M} & {\textit{Average-over-context}: average over word's encodings from $M$ different contexts/sentences}\\
        \cmidrule(lr){2-3}
        
        \multirow{3}{*}{\bf Subword Tokens} & {\nospec} & {Special tokens [CLS] and [SEP] are excluded from subword embedding averaging} \\
		& {\all} & Both special tokens [CLS] and [SEP] are included into subword embedding averaging \\
		& {\cls} & [CLS] is included into subword embedding averaging; [SEP] is excluded \\
        \cmidrule(lr){2-3}
		
		\multirow{2}{*}{\bf Layerwise Avg} & {\textsc{avg}(L$\leq$n)} & {Average representations over all Transformer layers up to the $n$-th layer $L_n$ (included)} \\
		& {L$=$n} & {Only the representation from the layer $L_{n}$ is used}\\
		
        \bottomrule
	\end{tabularx}}%
	\vspace{-1mm}
	\caption{Configuration components of word-level embedding extraction, resulting in 24 possible configurations. 
	}
	\label{tab:config}
    \vspace{-1mm}
\end{table*}

In this work, we explore several unsupervised word-level representation extraction strategies and configurations for lexico-semantic tasks (i.e., probes), stemming from different combinations of the components detailed in Table~\ref{tab:config} and illustrated in Figure~\ref{fig:configs}. In particular, we assess the impact of: \textbf{1)} encoding tokens with monolingual LM-pretrained Transformers vs.\,with their massively multilingual counterparts; \textbf{2)} providing context around the target word in input; \textbf{3)} including special tokens like [CLS] and [SEP]; \textbf{4)} averaging across several layers as opposed to a single layer.\footnote{For clarity of presentation, later in \S\ref{s:results} we show results only for a representative selection of configurations 
that are consistently better than the others} 



\section{Experimental Setup}
\label{s:experiments}
\noindent \textbf{Pretrained LMs and Languages.} 
Our selection of test languages is guided by the following constraints: \textbf{a)} availability of comparable pretrained (language-specific) monolingual LMs; \textbf{b)} availability of evaluation data; and \textbf{c)} typological diversity of the sample, along the lines of recent initiatives in multilingual NLP 
\cite[\textit{inter alia}]{Gerz2018on,hu2020xtreme,Ponti:2020xcopa}. 
We work with English (\en), German (\de), Russian (\ru), Finnish (\fin), Chinese (\zh), and Turkish (\tr). We use monolingual uncased BERT Base models for all languages, retrieved from the HuggingFace repository \cite{Wolf:2019hf}.\footnote{\url{https://huggingface.co/models}; the links to the actual BERT models are in the appendix.} All BERT models comprise 12 768-dimensional Transformer layers $\{L_1 \, \text{(bottom layer)}, \dots, L_{12} \, \text{(top)} \}$ plus the input embedding layer ($L_{0}$), and 12 attention heads. We also experiment with multilingual BERT (mBERT) \cite{devlin2018bert} as the underlying LM, aiming to measure the performance difference between language-specific and massively multilingual LMs  in our lexical probing tasks.


\vspace{1.8mm}
\noindent \textbf{Word Vocabularies and External Corpora.} We extract type-level representations in each language for the top 100K most frequent words represented in the respective fastText (FT) vectors, which were trained on lowercased monolingual Wikipedias by \newcite{Bojanowski:2017tacl}. The equivalent vocabulary coverage allows a direct comparison to fastText vectors, which we use as a baseline static WE method in all evaluation tasks. To retain the same vocabulary across all configurations, in \aoc variants we back off to the related \iso variant for words that have zero occurrences in external corpora.

For all \textsc{aoc} vector variants, we leverage 1M sentences of maximum sequence length 512, which we randomly sample from external corpora: Europarl \cite{Koehn:2005} for \en, \de, \fin, available via OPUS \cite{Tiedemann:2009opus}; the United Nations Parallel Corpus for \ru and \zh \cite{Ziemski:2016lrec}, and monolingual \tr WMT17 data \cite{Bojar:2017wmt}. 

\vspace{1.8mm}
\noindent \textbf{Evaluation Tasks.}
We carry out the evaluation on five standard and diverse lexical semantic tasks:

\vspace{1.4mm}
\noindent \textbf{Task 1: Lexical semantic similarity (LSIM)} is the most widespread intrinsic task for evaluation of traditional word embeddings \cite{Hill:2015cl}. The evaluation metric is the Spearman's rank correlation between the average of human-elicited semantic similarity scores for word pairs and the cosine similarity between the respective type-level word vectors. We rely on the recent comprehensive multilingual LSIM benchmark Multi-SimLex \cite{Vulic:2020msimlex}, which
covers 1,888 pairs in 13 languages. We focus on \en, \fin, \zh, \ru, the languages represented in Multi-SimLex.

\vspace{1.4mm}
\noindent \textbf{Task 2: Word Analogy (WA)} is another common intrinsic task. We evaluate our models on the Bigger Analogy Test Set (BATS) \cite{Drozd:2016coling} with 99,200 analogy questions.
We resort to the standard vector offset analogy resolution method, searching for the vocabulary word $w_d \in V$ such that its vector $d$ is obtained by $argmax_{d}(cos(d, c − a + b))$, where $a$, $b$, and $c$ are word vectors of words $w_a$, $w_b$, and $w_c$ from the analogy $w_a:w_b=w_c:x$. The search space comprises vectors of all words from the vocabulary $V$, excluding $a$, $b$, and $c$. 
This task is limited to \en, and we report Precision@1 scores.


\vspace{1.4mm}
\noindent \textbf{Task 3: Bilingual Lexicon Induction (BLI)} is a standard task to evaluate the ``semantic quality'' of static cross-lingual word embeddings (CLWEs) \cite{Gouws:2015icml,Ruder:2019jair}. We learn ``BERT-based'' CLWEs using a standard mapping-based approach \cite{Mikolov:2013arxiv,Smith:2017iclr} with \textsc{VecMap} \cite{Artetxe:2018acl}. BLI evaluation allows us to investigate the ``alignability'' of monolingual type-level representations extracted for different languages. We adopt the standard BLI evaluation setup  from \newcite{Glavas:2019acl}: 5K training word pairs are used to learn the mapping, and another 2K pairs as test data. We report standard Mean Reciprocal Rank (MRR) scores for 10 language pairs spanning \en, \de, \ru, \fin, \tr.
%

\vspace{1.4mm}
\noindent \textbf{Task 4: Cross-Lingual Information Retrieval (CLIR)}. We follow the setup of \newcite{Litschko:2018sigir,Litschko:2019sigir} and evaluate mapping-based CLWEs (the same ones as on BLI) in a document-level retrieval task on the CLEF 2003 benchmark.\footnote{All test collections comprise 60 queries. The average document collection size per language is 131K (ranging from 17K documents
for \ru to 295K for \de).} We use a simple CLIR model which showed competitive performance in the comparative studies of \newcite{Litschko:2019sigir} and \newcite{Glavas:2019acl}. It embeds queries and documents as IDF-weighted sums of their corresponding WEs from the CLWE space, and uses cosine similarity as the ranking function. We report Mean Average Precision (MAP) scores for 6 language pairs covering \en, \de, \ru, \fin.


\vspace{1.4mm}
\noindent \textbf{Task 5: Lexical Relation Prediction (RELP).} We probe if we can recover standard lexical relations (i.e., \textit{synonymy}, \textit{antonymy}, \textit{hypernymy}, \textit{meronymy}, plus \textit{no relation}) from input type-level vectors. We rely on a state-of-the-art neural model for RELP operating on type-level embeddings \cite{Glavas:2018naacl}: the Specialization Tensor Model (STM) predicts lexical relations for pairs of input word vectors based on multi-view projections of those vectors.\footnote{Note that RELP is structurally different from the other four tasks: instead of direct computations with word embeddings, called \textit{metric learning} or \textit{similarity-based} evaluation \cite{Ruder:2019jair}, it uses them as \textit{features} in a neural architecture.} We use the WordNet-based \cite{Fellbaum:1998wn} evaluation data of \newcite{Glavas:2018naacl}: they contain 10K annotated word pairs balanced by class.
Micro-averaged $F_1$ scores, averaged across 5 runs for each input vector space (default STM setting), are reported for \en and \de.

\begin{figure*}[p]
    \centering
    \begin{subfigure}[!ht]{0.88\textwidth}
        \centering
        \includegraphics[width=1.0\linewidth,trim=0cm 0cm 0cm 0.1cm,clip]{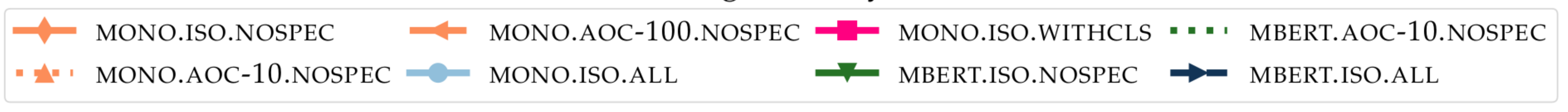}
        \label{fig:legend}
        \vspace{-3.5mm}
    \end{subfigure}
    \begin{subfigure}[!ht]{0.495\linewidth}
        \centering
        \includegraphics[width=0.88\linewidth]{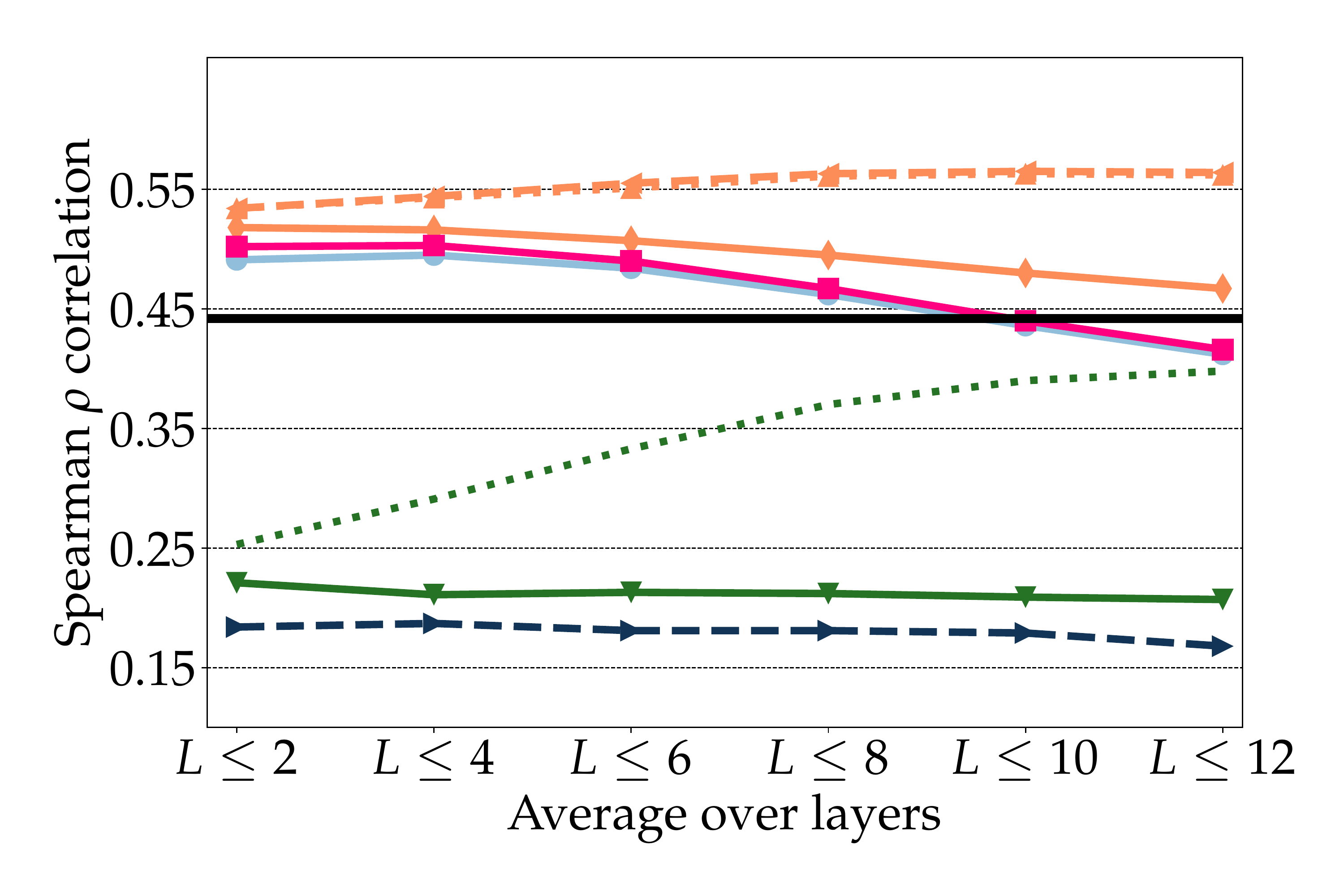}
        \caption{English}
        \label{fig:en}
    \end{subfigure}
    \begin{subfigure}[!ht]{0.495\textwidth}
        \centering
        \includegraphics[width=0.88\linewidth]{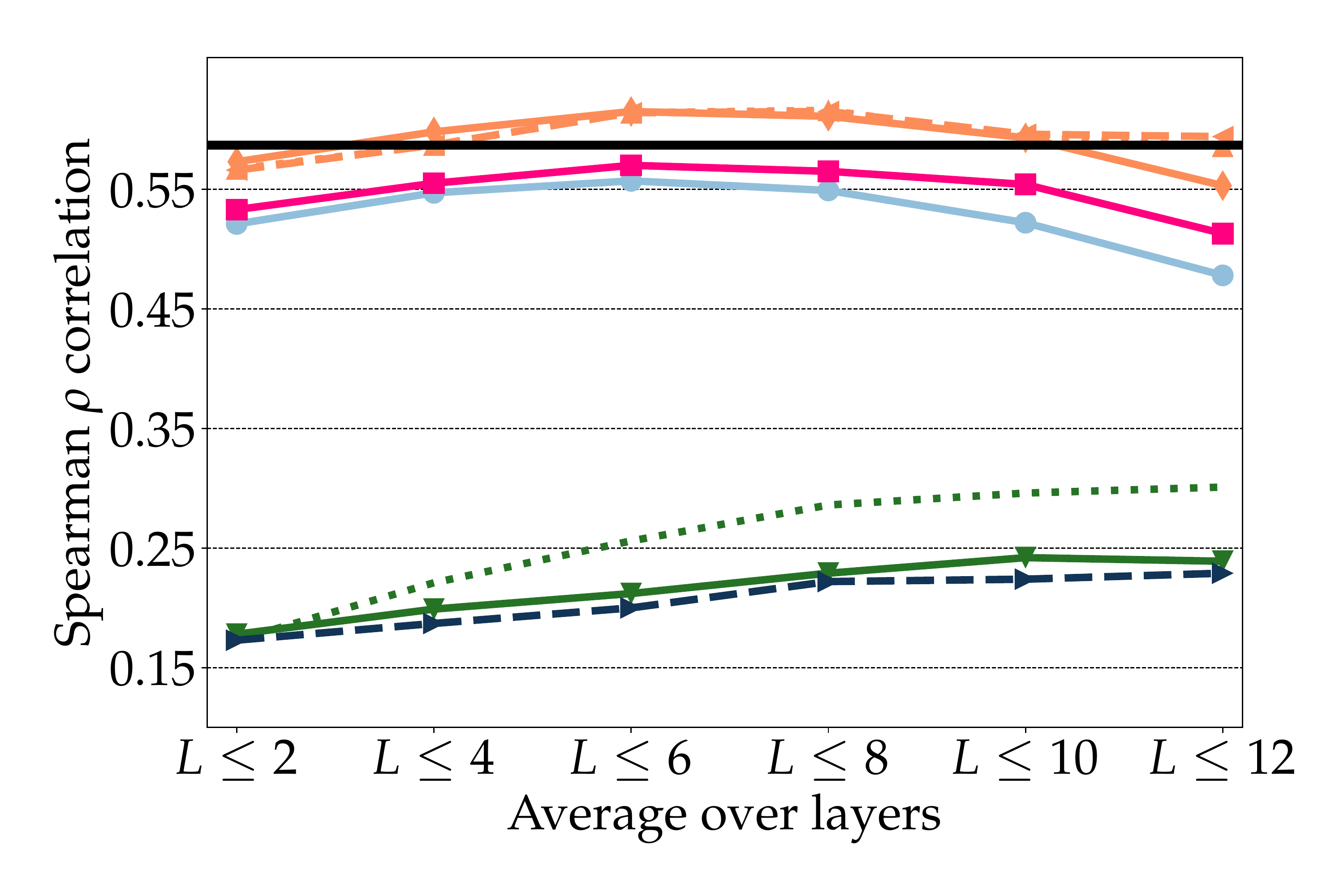}
        \caption{Finnish}
        \label{fig:fi}
    \end{subfigure}
    \begin{subfigure}[!ht]{0.495\linewidth}
        \centering
        \includegraphics[width=0.88\linewidth]{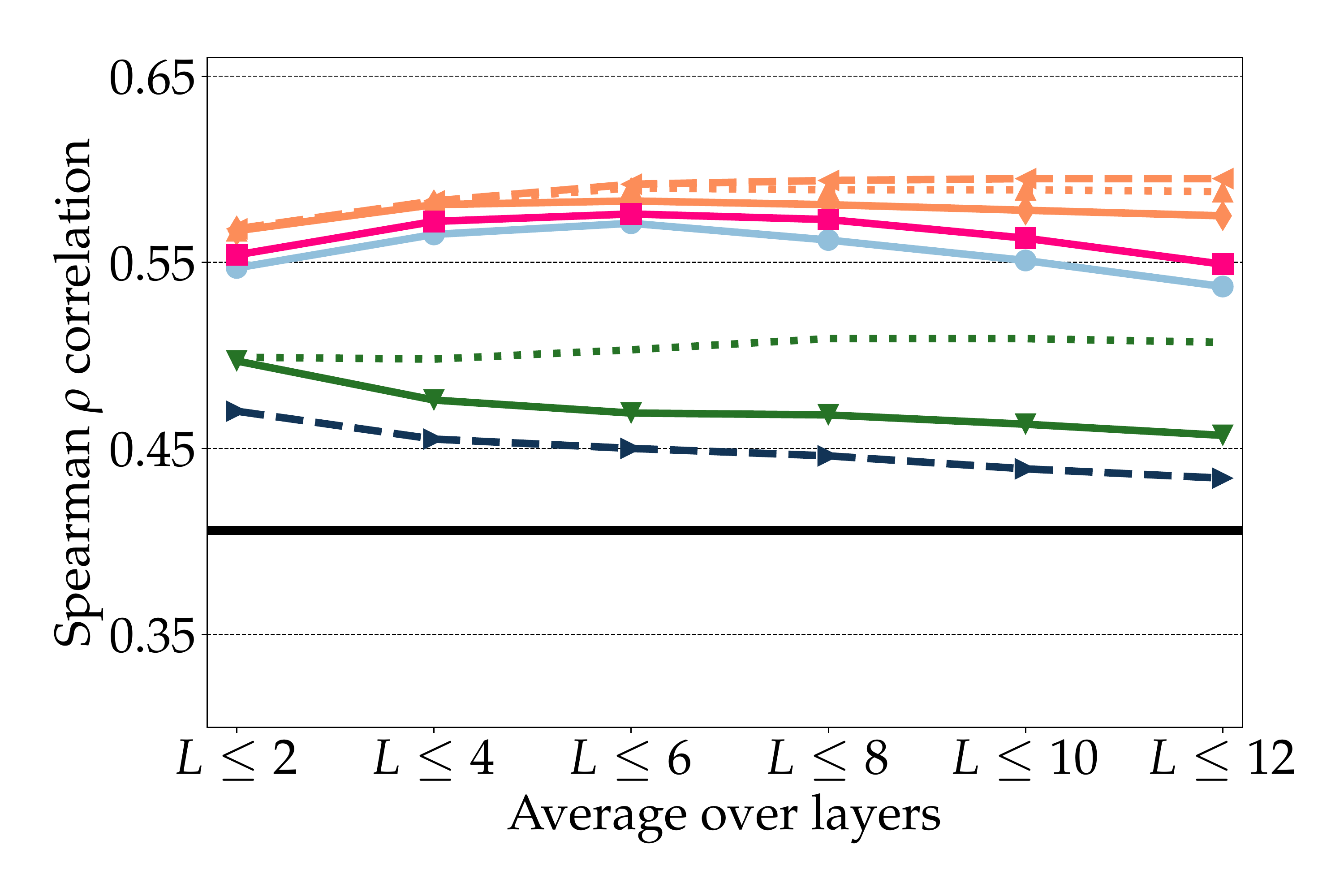}
        \caption{Mandarin Chinese}
        \label{fig:zh}
    \end{subfigure}
    \begin{subfigure}[!ht]{0.495\textwidth}
        \centering
        \includegraphics[width=0.88\linewidth]{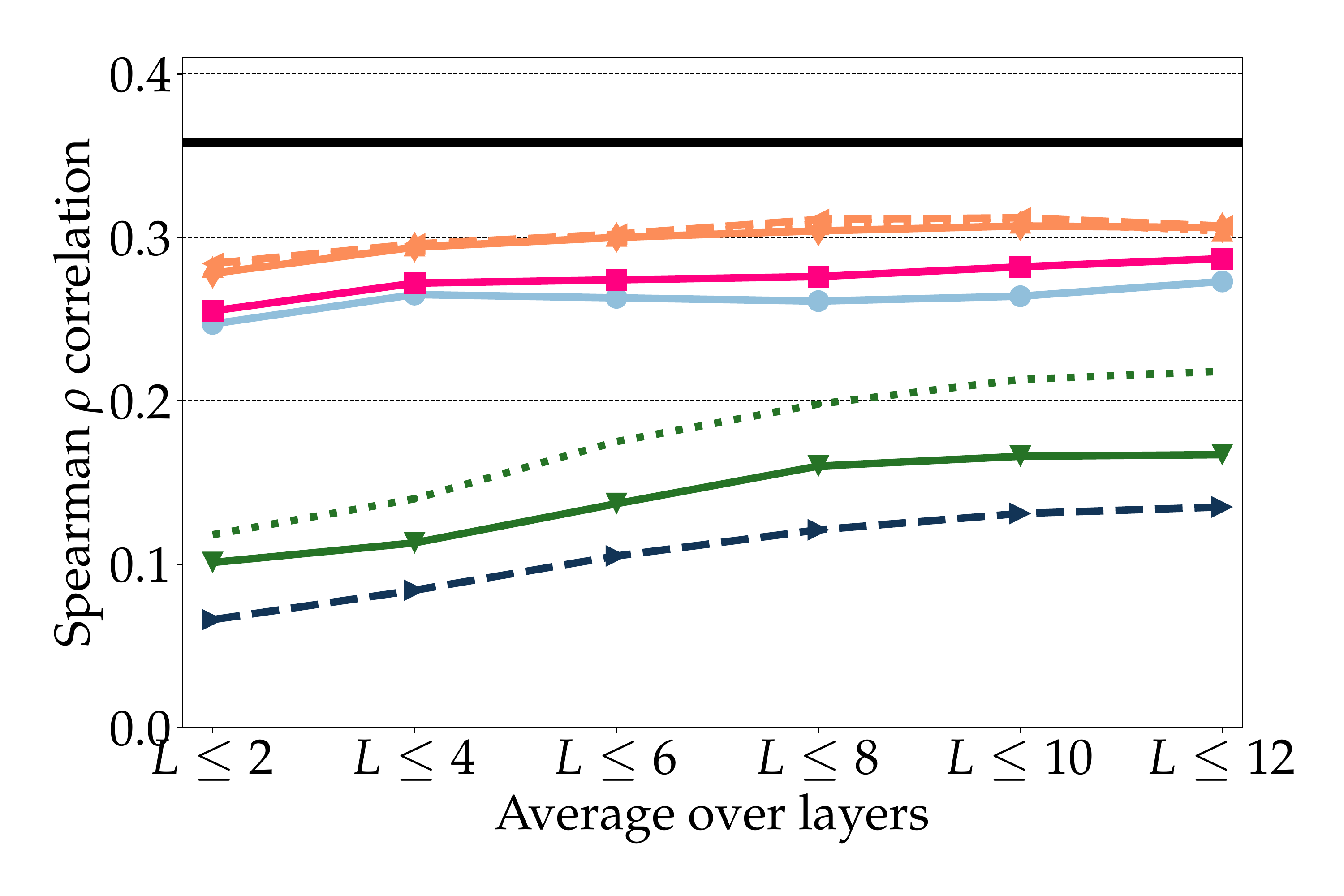}
        \caption{Russian}
        \label{fig:ru}
    \end{subfigure}
    \vspace{-2.5mm}
    \caption{Spearman's $\rho$ correlation scores for the lexical semantic similarity task (LSIM) in four languages. For the representation extraction configurations in the legend, see Table~\ref{tab:config}. Thick solid horizontal lines denote performance of standard monolingual fastText vectors trained on Wikipedia dumps of the respective languages.}
\label{fig:msimlex-summary}

    \centering
    \begin{subfigure}[!ht]{0.495\linewidth}
        \centering
        \includegraphics[width=0.88\linewidth]{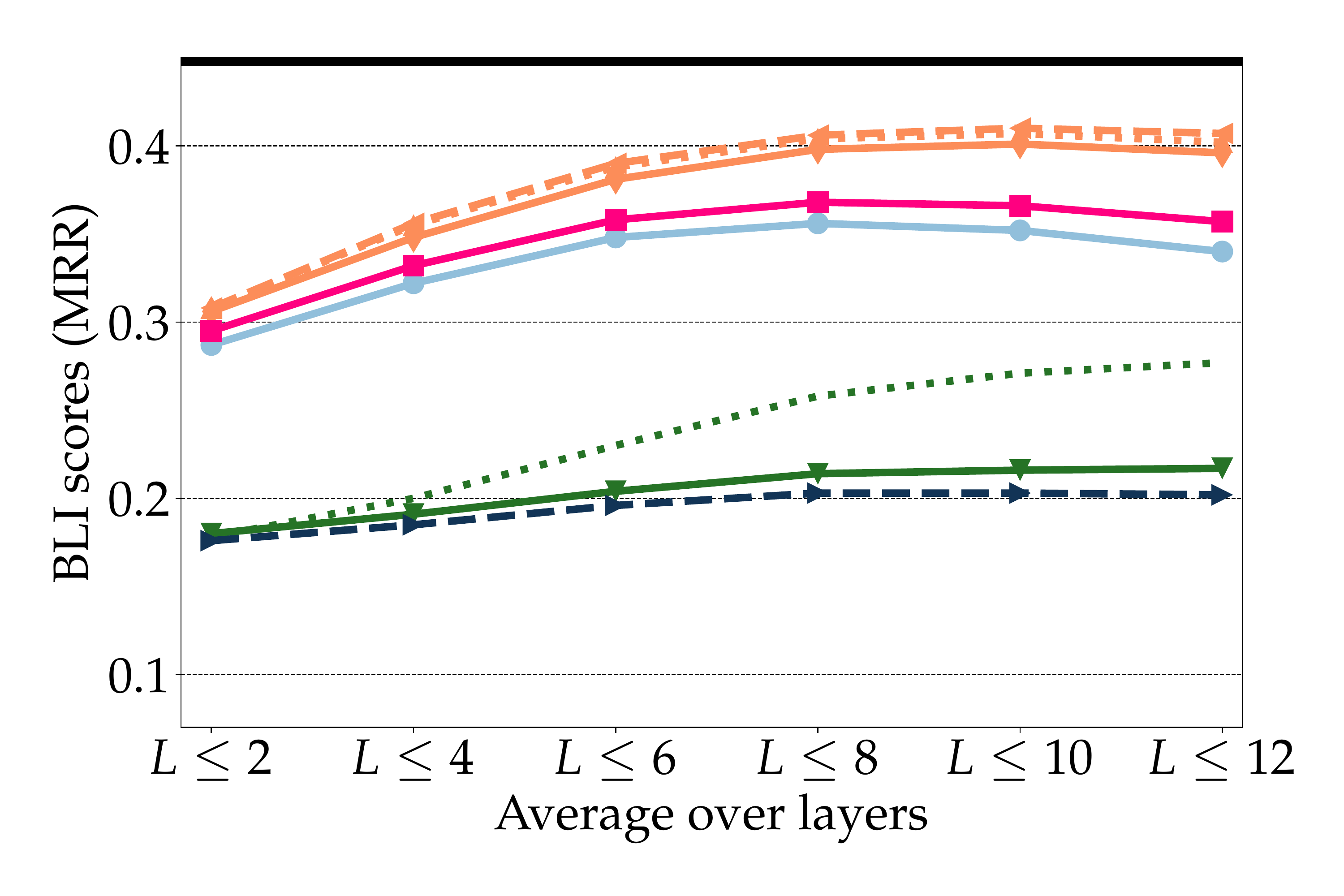}
        \caption{Summary BLI results}
        \label{fig:bli}
    \end{subfigure}
    \begin{subfigure}[!ht]{0.495\textwidth}
        \centering
        \includegraphics[width=0.88\linewidth]{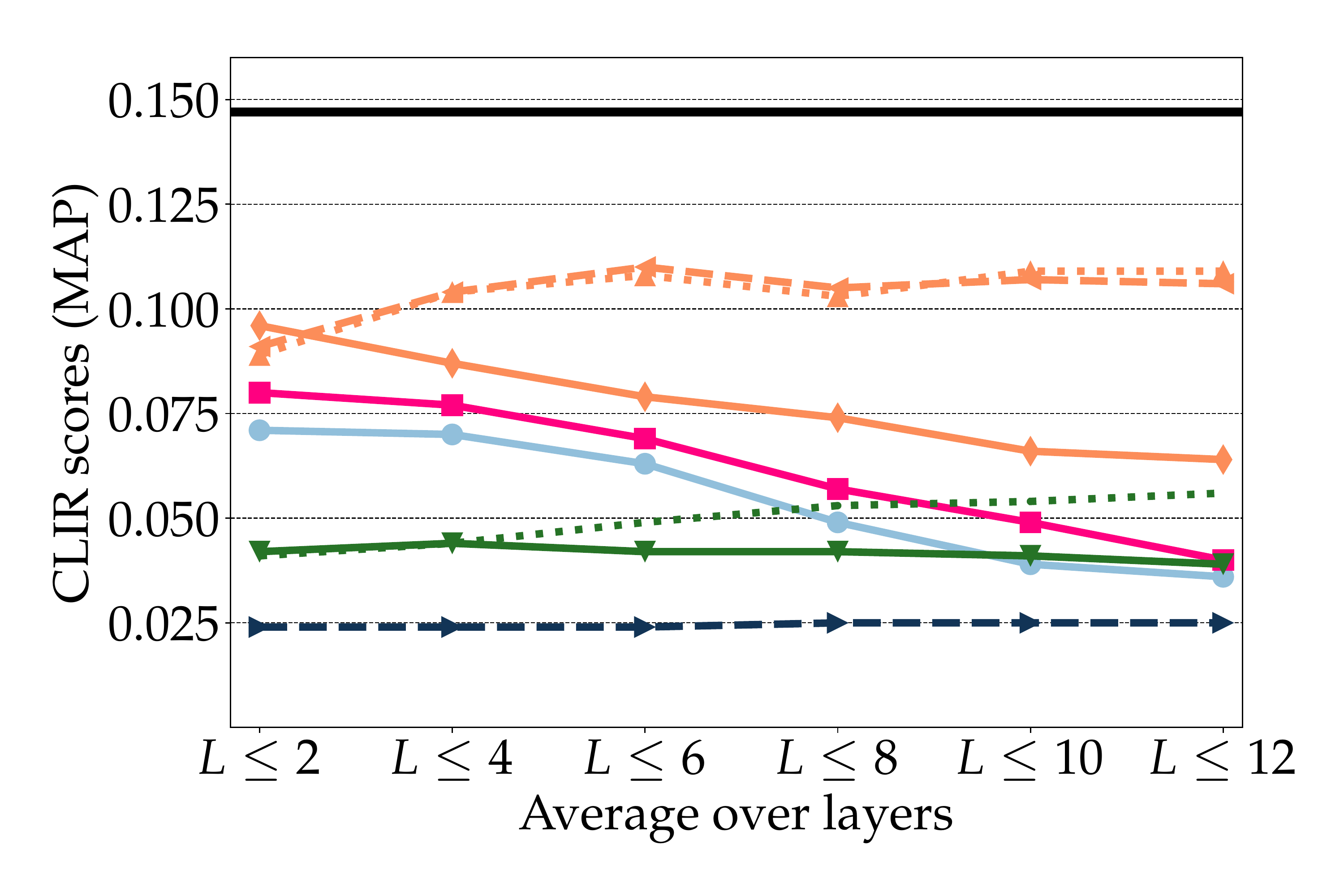}
        \caption{Summary CLIR results}
        \label{fig:clir}
    \end{subfigure}
    \vspace{-2.5mm}
    \caption{Summary results for the two cross-lingual evaluation tasks: \textbf{(a)} BLI (MRR scores) and \textbf{(b)} CLIR (MAP scores). We report average scores over all language pairs; 
    individual results for each language pair are available in the appendix. 
    Thick solid horizontal lines denote performance of standard fastText vectors in exactly the same cross-lingual mapping setup.}
\label{fig:bli-clir-summary}

    \centering
    \begin{subfigure}[!ht]{0.328\linewidth}
        \centering
        \includegraphics[width=0.99\linewidth]{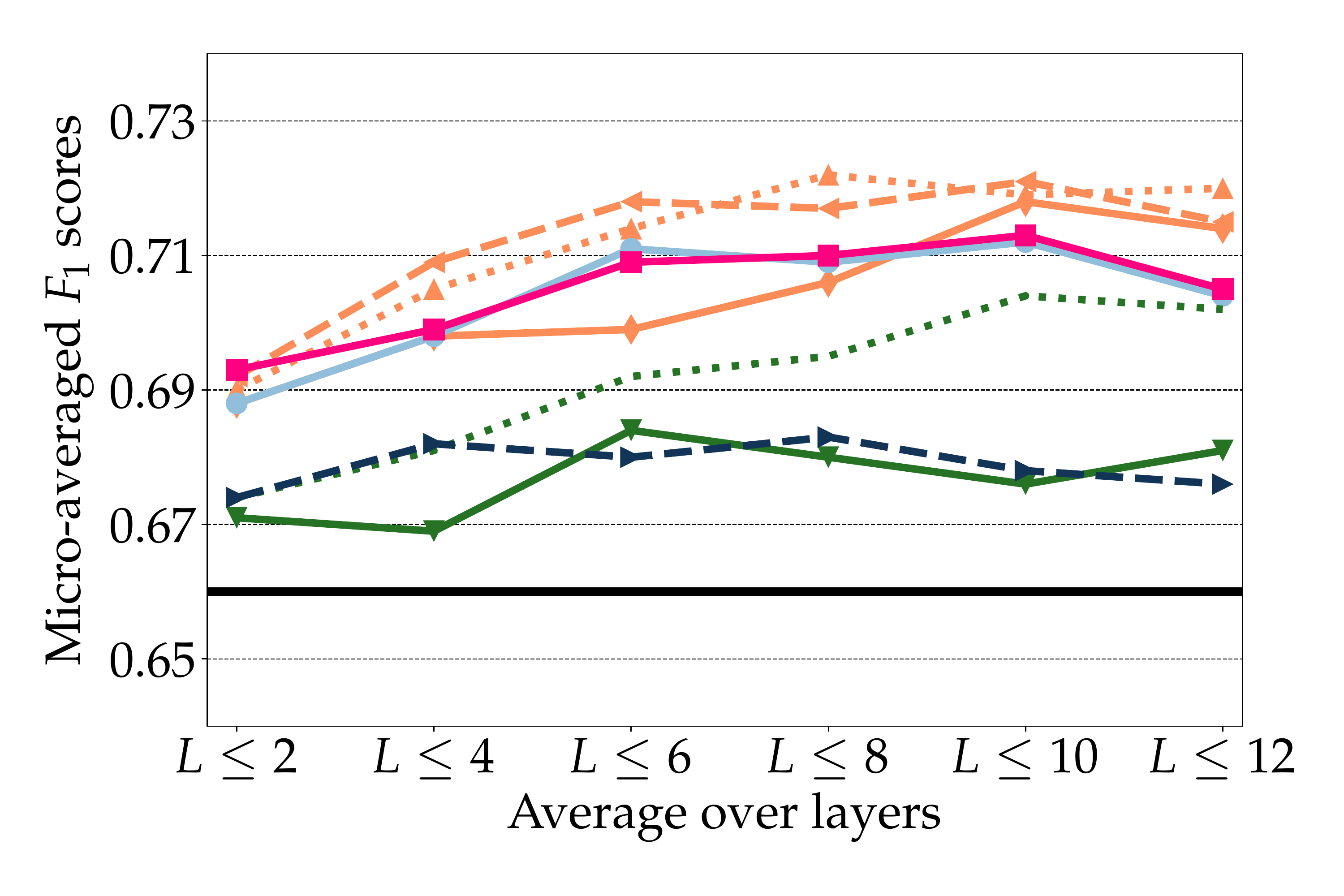}
        \caption{RELP: English}
        \label{fig:en-relp}
    \end{subfigure}
    \begin{subfigure}[!ht]{0.328\textwidth}
        \centering
        \includegraphics[width=0.99\linewidth]{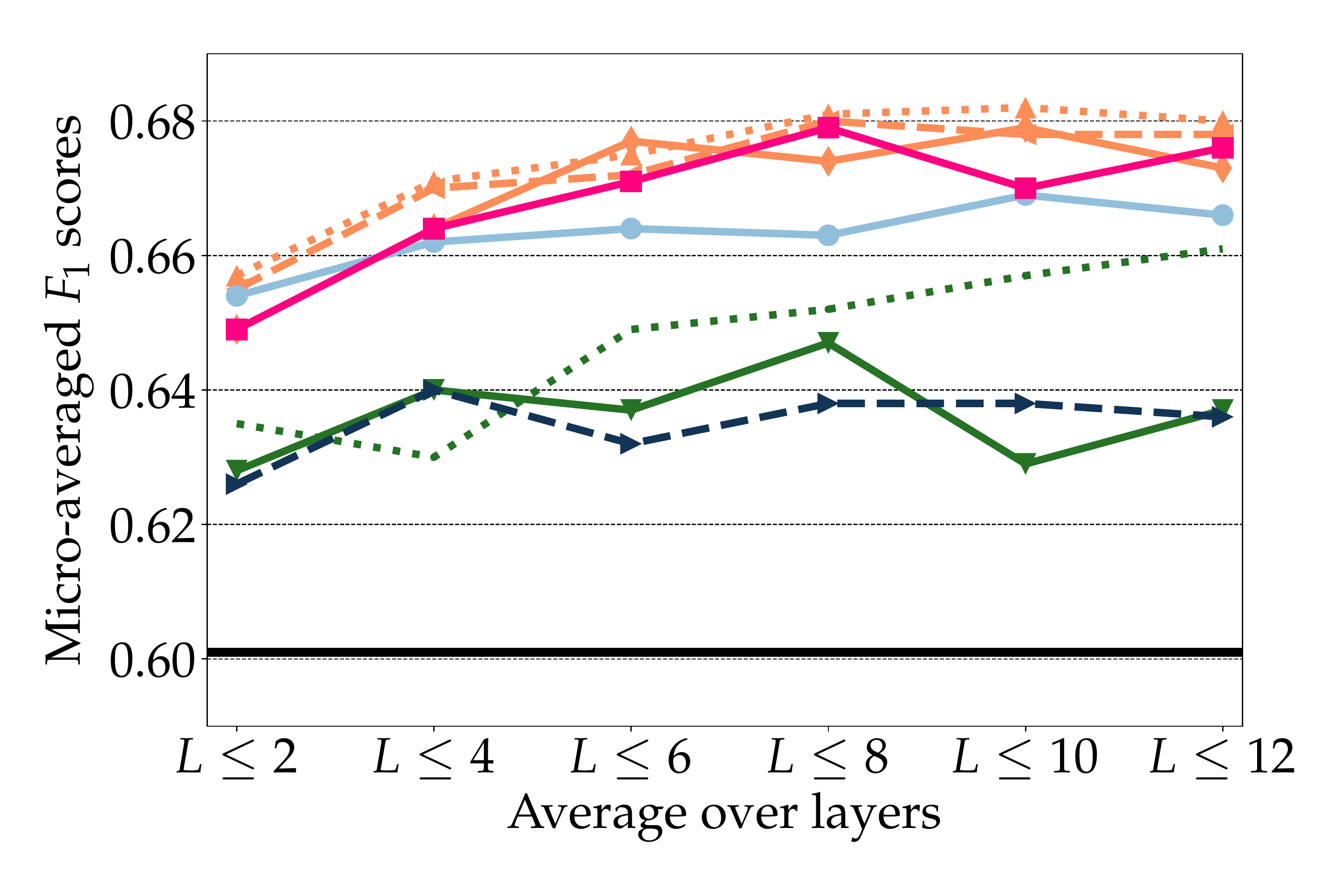}
        \caption{RELP: German}
        \label{fig:de-relp}
    \end{subfigure}
        \begin{subfigure}[!ht]{0.328\textwidth}
        \centering
        \includegraphics[width=0.99\linewidth]{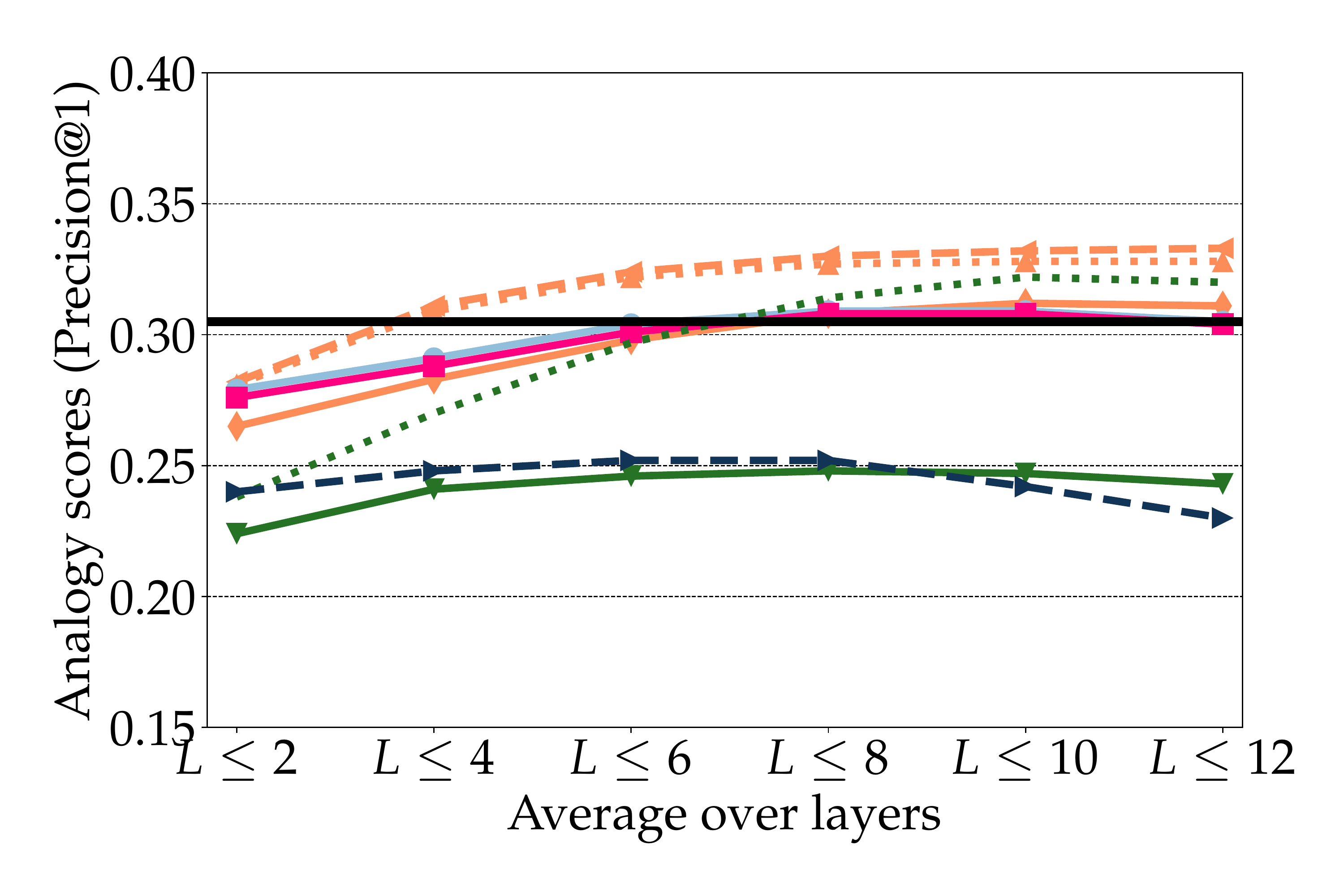}
        \caption{WA: English}
        \label{fig:en-wa}
    \end{subfigure}
    \vspace{-1.5mm}
    \caption{Micro-averaged $F_1$ scores in the RELP task for \textbf{(a)} \en and \textbf{(b)} \de. The scores  with 768-dim vectors randomly initalized via Xavier init \cite{Glorot:2010aistats} are 0.473 (\en) and 0.512 (\de); \textbf{(c)} \textsc{en} WA results.}
    \vspace{-2.5mm}
\label{fig:relp}
\end{figure*}

\section{Results and Discussion}
\label{s:results}


A summary of the results is shown in  Figure~\ref{fig:msimlex-summary} for LSIM, in Figure~\ref{fig:bli} for BLI, in Figure~\ref{fig:clir} for CLIR, in Figure~\ref{fig:en-relp} and Figure~\ref{fig:de-relp} for RELP, and in Figure~\ref{fig:en-wa} for WA. 
These results offer multiple axes of comparison, and the ensuing discussion focuses on the central questions Q1-Q3 posed in \S\ref{s:introduction}.\footnote{Full results are available in the appendix.}

\vspace{1.8mm}
\noindent \textbf{Monolingual versus Multilingual LMs.}
Results across all tasks validate the intuition that language-specific monolingual LMs contain much more lexical information for a particular target language than massively multilingual models such as mBERT or XLM-R \cite{Artetxe:2019arxiv}. We see large drops between \monobert.* and \mbert.* configurations even for very high-resource languages (\en and \de), and they are even more prominent for \fin and \tr. 

Encompassing 100+ training languages with limited model capacity, multilingual models suffer from the ``curse of multilinguality'' \cite{Conneau:2019arxiv}: they must trade off monolingual lexical information coverage (and consequently monolingual performance) for a wider language coverage.\footnote{For a particular target language, monolingual performance can be partially recovered by additional in-language monolingual training via masked language modeling \cite{Eisenschlos:2019emnlp,Pfeiffer:2020arxiv}. In a side experiment, we have also verified that the same holds for lexical information coverage. 
}

\vspace{1.8mm}
\noindent \textbf{How Important is Context?} 
Another observation that holds across all configurations concerns the usefulness of providing contexts drawn from external corpora, and corroborates findings from prior work \cite{Liu:2019conll}: \iso configurations cannot match configurations that average subword embeddings from multiple contexts (\aoc-\tens and \aoc-\huns). However, it is worth noting that \textbf{1)} performance gains with \aoc-100 over \aoc-10, although consistent, are quite marginal across all tasks: this suggests that several word occurrences \textit{in vivo} are already sufficient to accurately capture its type-level representation. \textbf{2)} In some tasks, \textsc{iso} configurations are only marginally outscored by their \aoc counterparts: e.g., for \monobert.*.\nospec.\textsc{avg}(L$\leq$8) on \en--\fin BLI or \de--\tr BLI, the respective scores are 0.486 and 0.315 with \iso, and 0.503 and 0.334 with \aoc-10. Similar observations hold for \fin and \zh LSIM, and also in the RELP task. 

In RELP, it is notable that `BERT-based' embeddings can recover more lexical relation knowledge than standard FT vectors. These findings reveal that pretrained LMs indeed implicitly capture plenty of lexical type-level knowledge (which needs to be `recovered' from the models); this also suggests why pretrained LMs have been successful in tasks where this knowledge is directly useful, such as NER and POS tagging \cite{Tenney:2019acl,Tsai:2019emnlp}. Finally, we also note that gains with \aoc over \iso are much more pronounced for the under-performing \mbert.* configurations: this indicates that \monobert models store more lexical information even in absence of context.


\vspace{1.8mm}
\noindent \textbf{How Important are Special Tokens?}
The results reveal that the inclusion of special tokens [CLS] and [SEP] into type-level embedding extraction deteriorates the final lexical information contained in the embeddings. This finding holds for different languages, underlying LMs, and averaging across various layers. The \textsc{nospec} configurations consistently outperform their \all and \cls counterparts, both in \iso and \aoc-\{10, 100\} settings.\footnote{For this reason, we report the results of \aoc configurations only in the \nospec setting.} 

Our finding at the lexical level aligns well with prior observations on using BERT directly as a sentence encoder \cite{Qiao:2019arxiv,Singh:2019ws,Casanueva:2020arxiv}: while [CLS] is useful for sentence-pair classification tasks, using [CLS] as a sentence representation produces inferior representations than averaging over sentence's subwords. In this work, we show that [CLS] and [SEP] should also be fully excluded from subword averaging for type-level word representations.


\vspace{1.8mm}
\noindent \textbf{How Important is Layer-wise Averaging?}
Averaging across layers bottom-to-top (i.e., from $L_0$ to $L_{12}$) is beneficial across the board, but we notice that scores typically saturate or even decrease in some tasks and languages when we include higher layers into averaging: see the scores with *.\textsc{avg}(L$\leq$10) and *.\textsc{avg}(L$\leq$12) configurations, e.g., for \fin LSIM; \en/\de RELP, and summary BLI and CLIR scores. This hints to the fact that two strategies typically used in prior work, either to take the vectors only from the embedding layer $L_0$ \cite{Wu:2019arxiv,Wang:2019emnlp} or to average across all layers \cite{Liu:2019conll}, extract sub-optimal word representations for a wide range of setups and languages. 

\begin{figure*}[!t]
    \centering
    \begin{subfigure}[!t]{0.328\linewidth}
        \centering
        \includegraphics[width=0.98\linewidth]{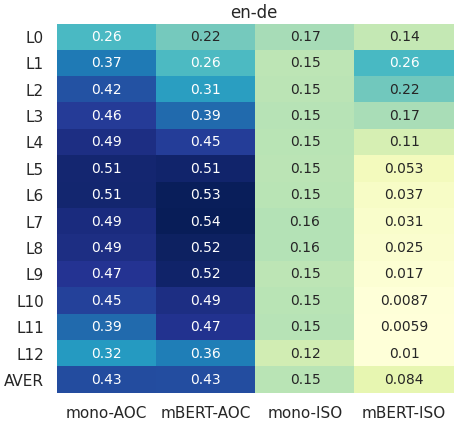}
        \label{fig:en-de-cka}
    \end{subfigure}
    \begin{subfigure}[!t]{0.328\textwidth}
        \centering
        \includegraphics[width=0.98\linewidth]{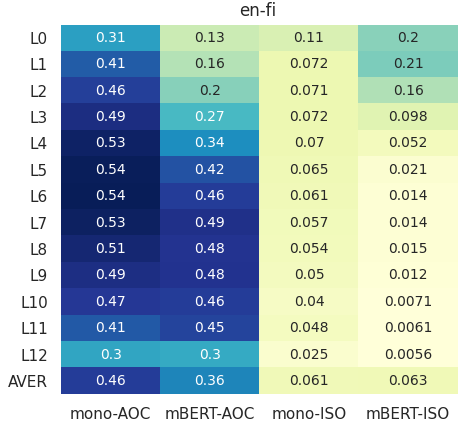}
        \label{fig:en-fi-cka}
    \end{subfigure}
        \begin{subfigure}[!t]{0.328\textwidth}
        \centering
        \includegraphics[width=0.98\linewidth]{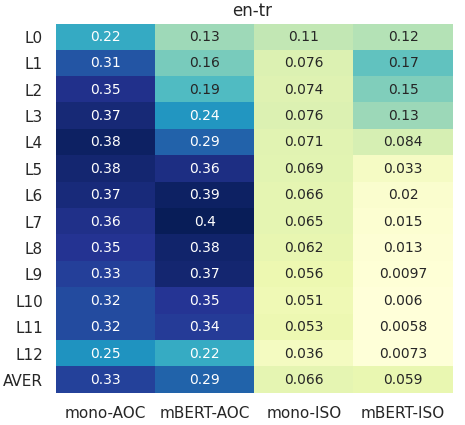}
        \label{fig:en-tr-cka}
    \end{subfigure}
    \vspace{-1mm}
    \caption{CKA similarity scores of type-level word representations extracted from each layer (using different extraction configurations, see Table~\ref{tab:config}) for a set of 7K translation pairs in \en--\de, \en--\fin, and \en--\tr from the BLI dictionaries of \newcite{Glavas:2019acl}. Additional heatmaps (where random words from two languages are paired) are available in the appendix.}
    \vspace{-1mm}
\label{fig:biling-cka}
\end{figure*}
\begin{figure*}[!t]
    \centering
    \begin{subfigure}[!ht]{0.488\linewidth}
        \centering
        \includegraphics[width=0.99\linewidth]{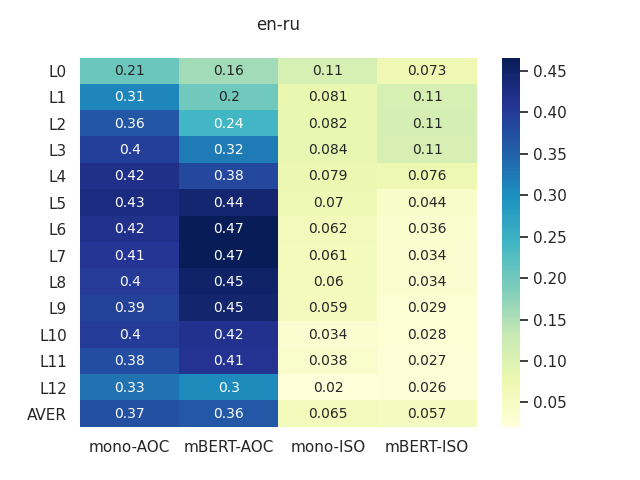}
        \caption{\en--\ru: Word translation pairs}
        \label{fig:en-ru}
    \end{subfigure}
    \begin{subfigure}[!ht]{0.488\textwidth}
        \centering
        \includegraphics[width=0.99\linewidth]{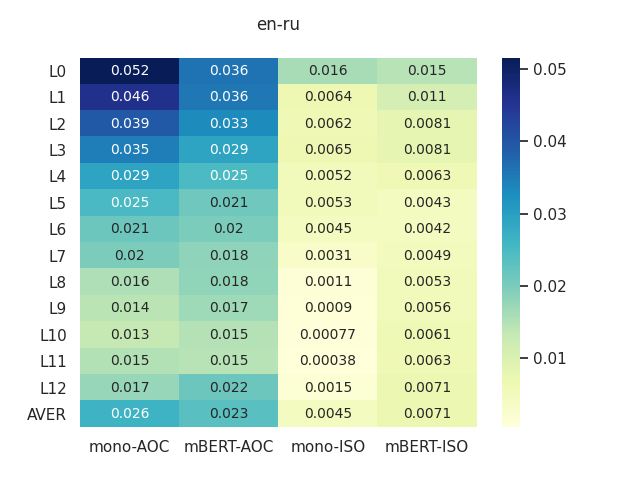}
        \caption{\en--\ru: Random word pairs}
        \label{fig:en-ru-random}
    \end{subfigure}
    \vspace{-1mm}
    \caption{CKA similarity scores of type-level word representations extracted from each layer for a set of \textbf{(a)} 7K \en--\ru translation pairs from the BLI dictionaries of \newcite{Glavas:2019acl}; \textbf{(b)} 7K random \en--\ru pairs.}
    \vspace{-1mm}
\label{fig:cka-ru}
\end{figure*}
\begin{figure*}[!t]
    \centering
    \begin{subfigure}[!t]{0.458\linewidth}
        \centering
        \includegraphics[width=0.96\linewidth]{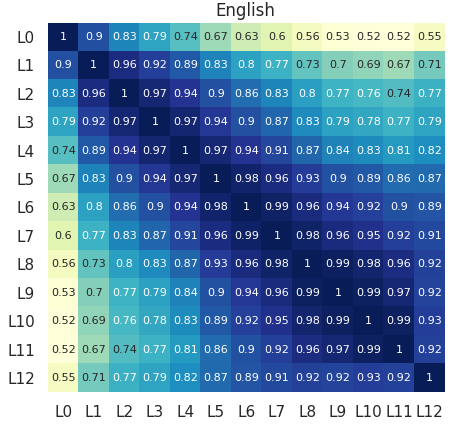}
        \label{fig:en-layer}
    \end{subfigure}
    \begin{subfigure}[!t]{0.458\textwidth}
        \centering
        \includegraphics[width=0.96\linewidth]{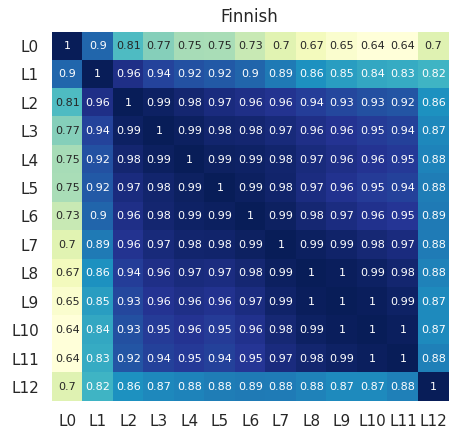}
        \label{fig:fi-layer}
    \end{subfigure}
    \vspace{-1mm}
    \caption{Self-similarity heatmaps: linear CKA similarity of representations for the same word extracted from different Transformer layers, averaged across 7K words for English and Finnish. \monobert.\aoc-100.\nospec.}
    \vspace{-1mm}
\label{fig:mono-cka}
\end{figure*}

\begin{table*}[!t]
\def\arraystretch{1.0}
\centering
{\footnotesize
\begin{adjustbox}{max width=\linewidth}
\begin{tabularx}{\textwidth}{ll YYYYYYYYYYYYY}
\toprule
& & {$L_0$} & {$L_1$} & {$L_2$} & {$L_3$} & {$L_4$} & {$L_5$} & {$L_6$} & {$L_7$} & {$L_8$} & {$L_9$} & {$L_{10}$} & {$L_{11}$} & {$L_{12}$} \\
\cmidrule(lr){3-15}
\multirow{2}{*}{LSIM} & {\en} & {.503} & {\bf .513} & {.505} & {.510} & {.505} & {.484} & {.459} & {.435} & {.402} & {.361} & {.362} & {.372} & {.390} \\
& {\fin} & {.445} & {\bf .466} & {.445} & {.436} & {.430} & {.434} & {.421} & {.404} & {.374} & {.346} & {.333} & {.324} & {.286} \\
\cmidrule(lr){3-15}
\multirow{1}{*}{WA} & {\en} & {.220} & {.272} & {\bf .293} & {.285} & {\bf .293} & {.261} & {.240} & {.217} & {.199} & {.171} & {.189} & {.221} & {.229} \\
\cmidrule(lr){3-15}
\multirow{3}{*}{BLI} & {\en--\de} & {.310} & {.354} & {.379} & {\bf .400} & {.394} & {.393} & {.373} & {.358} & {.311} & {.272} & {.273} & {.264} & {.287} \\
& {\en--\fin} & {.309} & {.339} & {.360} & {.367} & {\bf .369} & {.345} & {.329} & {.303} & {.279} & {.252} & {.231} & {.194} & {.192} \\
& {\de--\fin} & {.211} & {.245} & {.268} & {.283} & {.289} & {\bf .303} & {.291} & {.292} & {.288} & {.282} & {.262} & {.219} & {.236} \\
\cmidrule(lr){3-15}
\multirow{3}{*}{CLIR} & {\en--\de} & {.059} & {\bf .060} & {.059} & {\bf .060} & {.043} & {.036} & {.036} & {.036} & {.027} & {.024} & {.027} & {.035} & {.038} \\
& {\en--\fin} & {.038} & {\bf .040} & {.022} & {.018} & {.011} & {.008} & {.006} & {.006} & {.005} & {.002} & {.003} & {.002} & {.007} \\
& {\de--\fin} & {.054} & {\bf .057} & {.028} & {.015} & {.016} & {.022} & {.017} & {.021} & {.020} & {.023} & {.015} & {.008} & {.030} \\
\bottomrule
\end{tabularx}
\end{adjustbox}
}
\vspace{-1mm}
\caption{Task performance of word representations extracted from different Transformer layers for a selection of tasks, languages, and language pairs. Configuration: \monobert.\aoc-100.\nospec. Highest scores per row are in bold.}
\label{tab:per-layer}
\vspace{-1mm}
\end{table*}
\indent The sweet spot for $n$ in *.\textsc{avg}(L$\leq$n) configurations seems largely task- and language-dependent, as peak scores are obtained with different $n$-s. Whereas \textit{averaging across all layers} generally hurts performance, the results strongly suggest that averaging across layer subsets (rather than selecting a single layer) is widely useful, especially across bottom-most layers: e.g., $L\leq6$ with \monobert.\iso.\nospec yields an average score of 0.561 in LSIM, 0.076 in CLIR, and 0.432 in BLI; the respective scores when averaging over the 6 \textit{top} layers are: 0.218, 0.008, and 0.230. This evidence implies that, although scattered across multiple layers, type-level lexical information seems to be concentrated in lower Transformer layers. We investigate these conjectures further in \S\ref{ss:individual}.

\vspace{1.8mm}
\noindent \textbf{Comparison to Static Word Embeddings.}
The results also offer a comparison to static FT vectors across languages. The best-performing extraction configurations (e.g., \monobert.\aoc-\huns.\nospec) outperform FT in monolingual evaluations on LSIM (for \en, \fin, \zh), WA, and they also display much stronger performance in the RELP task for both evaluation languages. While the comparison is not strictly apples-to-apples, as FT and LMs were trained on different (Wikipedia) corpora, these findings leave open a provocative question for future work: \textit{Given that static type-level word representations can be recovered from large pretrained LMs, does this make standard static WEs obsolete, or are there applications where they are still useful?}

The trend is opposite in the two cross-lingual tasks: BLI and CLIR. While there are language pairs for which `BERT-based' WEs outperform FT (i.e., \en--\fin in BLI, \en--\ru and \fin--\ru in CLIR) or are very competitive to FT's performance (e.g., \en--\tr, \tr--\fi BLI, \de--\ru CLIR), FT provides higher scores overall in both tasks. The discrepancy between results in monolingual versus cross-lingual tasks warrants further investigation in future work. For instance, is using linear maps, as in standard mapping approaches to CLWE induction, sub-optimal for `BERT-based' word vectors?

\vspace{1.8mm}
\noindent \textbf{Differences across Languages and Tasks.} Finally, while we observe a conspicuous amount of universal patterns with configuration components (e.g., \monobert $>$ \mbert; \aoc $>$ \iso; \nospec $>$ \all, \cls), best-performing configurations do show some variation across different languages and tasks. For instance, while \en LSIM performance declines modestly but steadily when averaging over higher-level layers (\textsc{avg}(L$\leq n$), where $n>4$), performance on \en WA consistently increases for the same configurations. The BLI and CLIR scores in Figures~\ref{fig:bli} and \ref{fig:clir} also show slightly different patterns across layers. Overall, this suggests that \textbf{1)} extracted lexical information must be guided by task requirements, and \textbf{2)} config components must be carefully tuned to maximise performance for a particular task--language combination.


\subsection{Lexical Information in Individual Layers}
\label{ss:individual}
\noindent \textbf{Evaluation Setup.} To better understand which layers contribute the most to the final performance in our lexical tasks, we also probe type-level representations emerging from each individual layer of pretrained LMs. For brevity, we focus on the best performing configurations from previous experiments: \{\monobert, \textsc{mbert}\}.\{\iso, \aoc-100\}.\nospec.

In addition, tackling Q4 from \S\ref{s:introduction}, we analyse the similarity of representations extracted from  monolingual and multilingual BERT models using the centered kernel alignment (CKA) as proposed by \cite{Kornblith:2019icml}. The linear CKA computes similarity that is invariant to isotropic scaling and orthogonal transformation. It is defined as
%
{\normalsize
\begin{align}
\text{CKA}(X,Y)=\frac{\norm{Y^\top X}^2_\text{F}}{\left(\norm{X^\top X}_\text{F}\norm{Y^\top Y}_\text{F}\right)}.
\end{align}}%
\noindent $X,Y \in \mathbb{R}^{s \times d}$ are input matrices spanning $s$ $\ell_2$-normalized and mean-centered examples of dimensionality $d=768$. We use CKA in two different experiments: \textbf{1)} measuring \textit{self-similarity} where we compute CKA similarity of representations extracted from different layers for the same word; and \textbf{2)} measuring \textit{bilingual layer correspondence} where we compute CKA similarity of representations extracted from the same layer for two words constituting a translation pair. To this end, we again use BLI dictionaries of \newcite{Glavas:2019acl} (see \S\ref{s:experiments}) covering 7K pairs (training + test pairs).

\vspace{1.8mm}
\noindent \textbf{Discussion.}
Per-layer CKA similarities are provided in Figure~\ref{fig:mono-cka} (self-similarity) and  Figure~\ref{fig:biling-cka} (bilingual), and we show results of representations extracted from individual layers for selected evaluation setups and languages in Table~\ref{tab:per-layer}. We also plot bilingual layer correspondence of true word translations versus randomly paired words for \textsc{en}--\textsc{ru} in Figure~\ref{fig:cka-ru}. Figure~\ref{fig:mono-cka} reveals very similar patterns for both \en and \fin, and we also observe that self-similarity scores decrease for more distant layers (cf., similarity of $L_1$ and $L_2$ versus $L_1$ and $L_{12}$). However, despite structural similarities identified by linear CKA, the scores from Table~\ref{tab:per-layer} demonstrate that structurally similar layers might encode different amounts of lexical information: e.g., compare performance drops between $L_5$ and $L_8$ in all evaluation tasks.

The results in Table~\ref{tab:per-layer} further suggest that more type-level lexical information is available in lower layers, as all peak scores in the table are achieved with representations extracted from layers $L_1-L_5$. Much lower scores in type-level semantic tasks for higher layers also empirically validate a recent hypothesis of \newcite{Ethayarajh:2019emnlp} ``that contextualised word representations are more context-specific in higher layers.'' We also note that none of the results with L=$n$ configurations from Table~\ref{tab:config} can match best performing \textsc{avg}(L$\leq$\textit{n}) configurations with layer-wise averaging. This confirms our hypothesis that type-level lexical knowledge, although predominantly captured by lower layers, is disseminated across multiple layers, and layer-wise averaging is crucial to uncover that knowledge.


Further, Figure~\ref{fig:biling-cka} and Figure~\ref{fig:cka-ru} reveal that even LMs trained on monolingual data learn similar representations in corresponding layers for word translations (see the \monobert.\aoc columns). Intuitively, this similarity is much more pronounced with \aoc configurations with mBERT. The comparison of scores in Figure~\ref{fig:cka-ru} also reveals much higher correspondence scores for true translation pairs than for randomly paired words (i.e., the correspondence scores for random pairings are, as expected, random). Moreover, \mbert CKA similarity scores turn out to be higher for more similar language pairs (cf. \en--\de versus \en--\tr~\mbert.\aoc columns). This suggests that, similar to static WEs, type-level `BERT-based' WEs of different languages also display topological similarity, often termed \textit{approximate isomorphism} \cite{Sogaard:2018acl}, but its degree depends on language proximity. This also clarifies why representations extracted from two independently trained monolingual LMs can be linearly aligned, as validated by BLI and CLIR evaluation (Table~\ref{tab:per-layer} and Figure~\ref{fig:bli-clir-summary}).\footnote{Previous work has empirically validated that sentence representations for semantically similar inputs from different languages are less similar in higher Transformer layers \cite{Singh:2019ws,Wu:2019emnlp}. In Figure~\ref{fig:biling-cka}, we demonstrate that this is also the case for type-level lexical information; however, unlike sentence representations where highest similarity is reported in lowest layers, Figure~\ref{fig:biling-cka} suggests that highest CKA similarities are achieved in intermediate layers $L_5$-$L_8$.}

We also calculated the Spearman's correlation between CKA similarity scores for configurations \monobert.\aoc-\huns.\nospec.\textsc{avg}(L$\leq$\textit{n}), for all $n=0,\ldots,12$, and their  corresponding BLI scores on \en--\fin, \en--\de, and \de--\fin. The correlations are very high: $\rho=1.0, 0.83, 0.99$, respectively. This further confirms the approximate isomorphism hypothesis: it seems that higher structural similarities of representations extracted from monolingual pretrained LMs facilitate their cross-lingual alignment.







\section{Further Discussion and Conclusion}
\label{s:conclusion}
\noindent \textbf{What about Larger LMs and Corpora?}
Aspects of LM pretraining, such as the number of model parameters or the size of pretraining data, also impact lexical knowledge stored in the LM's parameters. Our preliminary experiments have verified that \en BERT-Large yields slight gains over the \en BERT-Base architecture used in our work (e.g., peak \en LSIM scores rise from 0.518 to 0.531). In a similar vein, we have run additional experiments with two available Italian (\ita) BERT-Base models with identical parameter setups, where one was trained on 13GB of \ita text, and the other on 81GB. In \en (BERT-Base)--\ita BLI and CLIR evaluations we measure improvements from 0.548 to 0.572 (BLI), and from 0.148 to 0.160 (CLIR) with the 81GB \ita model. In-depth analyses of these factors are out of the scope of this work, but they warrant further investigations. 

\noindent



\vspace{1.8mm}
\noindent \textbf{Opening Future Research Avenues.}
Our study has empirically validated that (monolingually) pretrained LMs store a wealth of type-level lexical knowledge, but effectively uncovering and extracting such knowledge from the LMs' parameters depends on several crucial components (see \S\ref{s:methodology}). In particular, some universal choices of configuration can be recommended: i) choosing monolingual LMs; ii) encoding words with multiple contexts; iii) excluding special tokens; iv) averaging over lower layers. Moreover, we found that type-level WEs extracted from pretrained LMs can surpass static WEs like fastText \citep{Bojanowski:2017tacl}.

This study has only scratched the surface of this research avenue. In future work, we plan to investigate how domains of external corpora affect \aoc configurations, and how to sample representative contexts from the corpora. We will also extend the study to more languages, more lexical semantic probes, and other larger underlying LMs. The difference in performance across layers 
also calls for more sophisticated lexical representation extraction methods (e.g., through layer weighting or attention) similar to meta-embedding approaches \cite{Yin:2016acl,Bollegala:2018coling,Kiela:2018emnlp}. Given the current large gaps between monolingual and multilingual LMs, we will also focus on lightweight methods to enrich lexical content in multilingual LMs \cite{Wang:2020kadapter,Pfeiffer:2020arxiv}.

\section*{Acknowledgments}
This work is supported by the ERC Consolidator Grant LEXICAL: Lexical Acquisition Across Languages (no 648909) awarded to Anna Korhonen. The work of Goran Glavaš and Robert Litschko is supported by the Baden-Württemberg Stiftung (AGREE grant of the Eliteprogramm). 

\bibliography{refs}

\begin{thebibliography}{66}
\expandafter\ifx\csname natexlab\endcsname\relax\def\natexlab#1{#1}\fi

\bibitem[{Artetxe et~al.(2018)Artetxe, Labaka, and Agirre}]{Artetxe:2018acl}
Mikel Artetxe, Gorka Labaka, and Eneko Agirre. 2018.
\newblock \href {http://aclweb.org/anthology/P18-1073} {A robust self-learning
  method for fully unsupervised cross-lingual mappings of word embeddings}.
\newblock In \emph{Proceedings of ACL}, pages 789--798.

\bibitem[{Artetxe et~al.(2020)Artetxe, Ruder, and Yogatama}]{Artetxe:2019arxiv}
Mikel Artetxe, Sebastian Ruder, and Dani Yogatama. 2020.
\newblock \href {http://arxiv.org/abs/1910.11856} {On the cross-lingual
  transferability of monolingual representations}.
\newblock In \emph{Proceedings of ACL}, pages 4623--4637.

\bibitem[{Belinkov and Glass(2019)}]{Belinkov:2019tacl}
Yonatan Belinkov and James~R. Glass. 2019.
\newblock \href {https://transacl.org/ojs/index.php/tacl/article/view/1570}
  {Analysis methods in neural language processing: {A} survey}.
\newblock \emph{Transactions of the Association of Computational Linguistics},
  7:49--72.

\bibitem[{Bengio et~al.(2003)Bengio, Ducharme, Vincent, and
  Jauvin}]{bengio2003neural}
Yoshua Bengio, R{\'e}jean Ducharme, Pascal Vincent, and Christian Jauvin. 2003.
\newblock \href {http://www.jmlr.org/papers/volume3/bengio03a/bengio03a.pdf} {A
  neural probabilistic language model}.
\newblock \emph{Journal of Machine Learning Research}, 3:1137--1155.

\bibitem[{Bojanowski et~al.(2017)Bojanowski, Grave, Joulin, and
  Mikolov}]{Bojanowski:2017tacl}
Piotr Bojanowski, Edouard Grave, Armand Joulin, and Tomas Mikolov. 2017.
\newblock \href {http://arxiv.org/abs/1607.04606} {Enriching word vectors with
  subword information}.
\newblock \emph{Transactions of the ACL}, 5:135--146.

\bibitem[{Bojar et~al.(2017)Bojar, Chatterjee, Federmann, Graham, Haddow,
  Huang, Huck, Koehn, Liu, Logacheva, Monz, Negri, Post, Rubino, Specia, and
  Turchi}]{Bojar:2017wmt}
Ond{\v{r}}ej Bojar, Rajen Chatterjee, Christian Federmann, Yvette Graham, Barry
  Haddow, Shujian Huang, Matthias Huck, Philipp Koehn, Qun Liu, Varvara
  Logacheva, Christof Monz, Matteo Negri, Matt Post, Raphael Rubino, Lucia
  Specia, and Marco Turchi. 2017.
\newblock \href {https://www.aclweb.org/anthology/W17-4717} {Findings of the
  2017 {Conference on Machine Translation} ({WMT}17)}.
\newblock In \emph{Proceedings of WMT}, pages 169--214.

\bibitem[{Bollegala and Bao(2018)}]{Bollegala:2018coling}
Danushka Bollegala and Cong Bao. 2018.
\newblock \href {https://www.aclweb.org/anthology/C18-1140/} {Learning word
  meta-embeddings by autoencoding}.
\newblock In \emph{Proceedings of COLING}, pages 1650--1661.

\bibitem[{Casanueva et~al.(2020)Casanueva, Tem\v{c}inas, Gerz, Henderson, and
  Vuli\'{c}}]{Casanueva:2020arxiv}
I{\~{n}}igo Casanueva, Tadas Tem\v{c}inas, Daniela Gerz, Matthew Henderson, and
  Ivan Vuli\'{c}. 2020.
\newblock \href {https://arxiv.org/abs/2003.04807} {Efficient intent detection
  with dual sentence encoders}.
\newblock In \emph{Proceedings of the 2nd Workshop on Natural Language
  Processing for Conversational AI}, pages 38--45.

\bibitem[{Chang and Chen(2019)}]{chang-chen-2019-word}
Ting-Yun Chang and Yun-Nung Chen. 2019.
\newblock \href {https://www.aclweb.org/anthology/D19-1627} {What does this
  word mean? {E}xplaining contextualized embeddings with natural language
  definition}.
\newblock In \emph{Proceedings of EMNLP-IJCNLP}, pages 6064--6070.

\bibitem[{Chi et~al.(2020)Chi, Hewitt, and Manning}]{Chi:2020acl}
Ethan~A. Chi, John Hewitt, and Christopher~D. Manning. 2020.
\newblock \href {https://arxiv.org/abs/2005.04511} {Finding universal
  grammatical relations in multilingual {BERT}}.
\newblock In \emph{Proceedings of ACL}, pages 5564--5577.

\bibitem[{Conneau et~al.(2020)Conneau, Khandelwal, Goyal, Chaudhary, Wenzek,
  Guzm{\'{a}}n, Grave, Ott, Zettlemoyer, and Stoyanov}]{Conneau:2019arxiv}
Alexis Conneau, Kartikay Khandelwal, Naman Goyal, Vishrav Chaudhary, Guillaume
  Wenzek, Francisco Guzm{\'{a}}n, Edouard Grave, Myle Ott, Luke Zettlemoyer,
  and Veselin Stoyanov. 2020.
\newblock \href {http://arxiv.org/abs/1911.02116} {Unsupervised cross-lingual
  representation learning at scale}.
\newblock In \emph{Proceedings of ACL}, pages 8440--8451.

\bibitem[{Devlin et~al.(2019)Devlin, Chang, Lee, and
  Toutanova}]{devlin2018bert}
Jacob Devlin, Ming-Wei Chang, Kenton Lee, and Kristina Toutanova. 2019.
\newblock \href {https://www.aclweb.org/anthology/N19-1423} {{BERT}:
  Pre-training of deep bidirectional transformers for language understanding}.
\newblock In \emph{Proceedings of NAACL-HLT}, pages 4171--4186.

\bibitem[{Drozd et~al.(2016)Drozd, Gladkova, and Matsuoka}]{Drozd:2016coling}
Aleksandr Drozd, Anna Gladkova, and Satoshi Matsuoka. 2016.
\newblock \href {https://www.aclweb.org/anthology/C16-1332/} {Word embeddings,
  analogies, and machine learning: {B}eyond king - man + woman = queen}.
\newblock In \emph{Proceedings of COLING}, pages 3519--3530.

\bibitem[{Edmiston(2020)}]{Edmiston:2020arxiv}
Daniel Edmiston. 2020.
\newblock \href {https://arxiv.org/abs/2004.03032} {A systematic analysis of
  morphological content in {BERT} models for multiple languages}.
\newblock \emph{CoRR}, abs/2004.03032.

\bibitem[{Eisenschlos et~al.(2019)Eisenschlos, Ruder, Czapla, Kardas, Gugger,
  and Howard}]{Eisenschlos:2019emnlp}
Julian Eisenschlos, Sebastian Ruder, Piotr Czapla, Marcin Kardas, Sylvain
  Gugger, and Jeremy Howard. 2019.
\newblock \href {https://doi.org/10.18653/v1/D19-1572} {{MultiFiT: E}fficient
  multi-lingual language model fine-tuning}.
\newblock In \emph{Proceedings of EMNLP-IJCNLP}, pages 5701--5706.

\bibitem[{Ethayarajh(2019)}]{Ethayarajh:2019emnlp}
Kawin Ethayarajh. 2019.
\newblock \href {https://www.aclweb.org/anthology/D19-1006} {How contextual are
  contextualized word representations? {C}omparing the geometry of {BERT},
  {ELM}o, and {GPT}-2 embeddings}.
\newblock In \emph{Proceedings of EMNLP-IJCNLP}, pages 55--65.

\bibitem[{Fellbaum(1998)}]{Fellbaum:1998wn}
Christiane Fellbaum. 1998.
\newblock \href {https://mitpress.mit.edu/books/wordnet} {\emph{WordNet}}.
\newblock {}MIT Press.

\bibitem[{Gerz et~al.(2018)Gerz, Vuli{\'c}, Ponti, Reichart, and
  Korhonen}]{Gerz2018on}
Daniela Gerz, Ivan Vuli{\'c}, Edoardo~Maria Ponti, Roi Reichart, and Anna
  Korhonen. 2018.
\newblock \href {https://www.aclweb.org/anthology/D18-1029} {On the relation
  between linguistic typology and (limitations of) multilingual language
  modeling}.
\newblock In \emph{Proceedings of EMNLP}, pages 316--327.

\bibitem[{Glava{\v{s}} and Vuli{\'c}(2018)}]{Glavas:2018naacl}
Goran Glava{\v{s}} and Ivan Vuli{\'c}. 2018.
\newblock \href {https://www.aclweb.org/anthology/N18-2029} {Discriminating
  between lexico-semantic relations with the specialization tensor model}.
\newblock In \emph{Proceedings of NAACL-HLT}, pages 181--187.

\bibitem[{Glava\v{s} et~al.(2019)Glava\v{s}, Litschko, Ruder, and
  Vuli\'{c}}]{Glavas:2019acl}
Goran Glava\v{s}, Robert Litschko, Sebastian Ruder, and Ivan Vuli\'{c}. 2019.
\newblock \href {https://arxiv.org/pdf/1902.00508.pdf} {How to (properly)
  evaluate cross-lingual word embeddings: On strong baselines, comparative
  analyses, and some misconceptions}.
\newblock In \emph{Proceedings of ACL}, pages 710--721.

\bibitem[{Glorot and Bengio(2010)}]{Glorot:2010aistats}
Xavier Glorot and Yoshua Bengio. 2010.
\newblock \href {http://proceedings.mlr.press/v9/glorot10a/glorot10a.pdf}
  {Understanding the difficulty of training deep feedforward neural networks}.
\newblock In \emph{Proceedings of AISTATS}, pages 249--256.

\bibitem[{Gouws et~al.(2015)Gouws, Bengio, and Corrado}]{Gouws:2015icml}
Stephan Gouws, Yoshua Bengio, and Greg Corrado. 2015.
\newblock \href {http://proceedings.mlr.press/v37/gouws15.pdf} {{BilBOWA: F}ast
  bilingual distributed representations without word alignments}.
\newblock In \emph{Proceedings of ICML}, pages 748--756.

\bibitem[{Hewitt and Manning(2019)}]{Hewitt:2019naacl}
John Hewitt and Christopher~D. Manning. 2019.
\newblock \href {https://doi.org/10.18653/v1/n19-1419} {A structural probe for
  finding syntax in word representations}.
\newblock In \emph{Proceedings of NAACL-HLT}, pages 4129--4138.

\bibitem[{Hill et~al.(2015)Hill, Reichart, and Korhonen}]{Hill:2015cl}
Felix Hill, Roi Reichart, and Anna Korhonen. 2015.
\newblock \href {https://doi.org/10.1162/COLI\_a\_00237} {{SimLex-999:
  E}valuating semantic models with (genuine) similarity estimation}.
\newblock \emph{Computational Linguistics}, 41(4):665--695.

\bibitem[{Hofmann et~al.(2020)Hofmann, Pierrehumbert, and
  Sch{\"{u}}tze}]{Hofmann:2020arxiv}
Valentin Hofmann, Janet~B. Pierrehumbert, and Hinrich Sch{\"{u}}tze. 2020.
\newblock \href {https://arxiv.org/abs/2005.00672} {Generating derivational
  morphology with {BERT}}.
\newblock \emph{CoRR}, abs/2005.00672.

\bibitem[{Hu et~al.(2020)Hu, Ruder, Siddhant, Neubig, Firat, and
  Johnson}]{hu2020xtreme}
Junjie Hu, Sebastian Ruder, Aditya Siddhant, Graham Neubig, Orhan Firat, and
  Melvin Johnson. 2020.
\newblock \href {https://arxiv.org/pdf/2003.11080.pdf} {{XTREME}: A massively
  multilingual multi-task benchmark for evaluating cross-lingual
  generalization}.
\newblock In \emph{Proceedings of ICML}.

\bibitem[{Jawahar et~al.(2019)Jawahar, Sagot, and Seddah}]{Jawahar:2019acl}
Ganesh Jawahar, Beno{\^{\i}}t Sagot, and Djam{\'{e}} Seddah. 2019.
\newblock \href {https://doi.org/10.18653/v1/p19-1356} {What does {BERT} learn
  about the structure of language?}
\newblock In \emph{Proceedings of ACL}, pages 3651--3657.

\bibitem[{Kiela et~al.(2018)Kiela, Wang, and Cho}]{Kiela:2018emnlp}
Douwe Kiela, Changhan Wang, and Kyunghyun Cho. 2018.
\newblock \href {https://doi.org/10.18653/v1/d18-1176} {Dynamic meta-embeddings
  for improved sentence representations}.
\newblock In \emph{Proceedings of EMNLP}, pages 1466--1477.

\bibitem[{Koehn(2005)}]{Koehn:2005}
Philipp Koehn. 2005.
\newblock \href {http://www.statmt.org/europarl/} {Europarl: {A} parallel
  corpus for statistical machine translation}.
\newblock In \emph{{Proceedings of the 10th Machine Translation Summit (MT
  SUMMIT)}}, pages 79--86.

\bibitem[{Kornblith et~al.(2019)Kornblith, Norouzi, Lee, and
  Hinton}]{Kornblith:2019icml}
Simon Kornblith, Mohammad Norouzi, Honglak Lee, and Geoffrey~E. Hinton. 2019.
\newblock \href {http://proceedings.mlr.press/v97/kornblith19a.html}
  {Similarity of neural network representations revisited}.
\newblock In \emph{Proceedings of ICML}, pages 3519--3529.

\bibitem[{Kulmizev et~al.(2020)Kulmizev, Ravishankar, Abdou, and
  Nivre}]{Kulmizev:2020acl}
Artur Kulmizev, Vinit Ravishankar, Mostafa Abdou, and Joakim Nivre. 2020.
\newblock \href {https://arxiv.org/abs/2004.14096} {Do neural language models
  show preferences for syntactic formalisms?}
\newblock In \emph{Proceedings of ACL}, pages 4077--4091.

\bibitem[{Litschko et~al.(2018)Litschko, Glava\v{s}, Ponzetto, and
  Vuli\'{c}}]{Litschko:2018sigir}
Robert Litschko, Goran Glava\v{s}, Simone~Paolo Ponzetto, and Ivan Vuli\'{c}.
  2018.
\newblock \href {https://arxiv.org/abs/1805.00879} {Unsupervised cross-lingual
  information retrieval using monolingual data only}.
\newblock In \emph{Proceedings of SIGIR}, pages 1253--1256.

\bibitem[{Litschko et~al.(2019)Litschko, Glava\v{s}, Vuli\'{c}, and
  Dietz}]{Litschko:2019sigir}
Robert Litschko, Goran Glava\v{s}, Ivan Vuli\'{c}, and Laura Dietz. 2019.
\newblock \href {https://doi.org/10.1145/3331184.3331324} {Evaluating
  resource-lean cross-lingual embedding models in unsupervised retrieval}.
\newblock In \emph{Proceedings of SIGIR}, pages 1109--1112.

\bibitem[{Liu et~al.(2019{\natexlab{a}})Liu, Gardner, Belinkov, Peters, and
  Smith}]{Liu:2019naacl}
Nelson~F. Liu, Matt Gardner, Yonatan Belinkov, Matthew~E. Peters, and Noah~A.
  Smith. 2019{\natexlab{a}}.
\newblock \href {https://www.aclweb.org/anthology/N19-1112} {Linguistic
  knowledge and transferability of contextual representations}.
\newblock In \emph{Proceedings of NAACL-HLT}, pages 1073--1094.

\bibitem[{Liu et~al.(2019{\natexlab{b}})Liu, McCarthy, Vuli{\'c}, and
  Korhonen}]{Liu:2019conll}
Qianchu Liu, Diana McCarthy, Ivan Vuli{\'c}, and Anna Korhonen.
  2019{\natexlab{b}}.
\newblock \href {https://www.aclweb.org/anthology/K19-1004} {Investigating
  cross-lingual alignment methods for contextualized embeddings with
  token-level evaluation}.
\newblock In \emph{Proceedings of CoNLL}, pages 33--43.

\bibitem[{Liu et~al.(2019{\natexlab{c}})Liu, Ott, Goyal, Du, Joshi, Chen, Levy,
  Lewis, Zettlemoyer, and Stoyanov}]{Liu:2019roberta}
Yinhan Liu, Myle Ott, Naman Goyal, Jingfei Du, Mandar Joshi, Danqi Chen, Omer
  Levy, Mike Lewis, Luke Zettlemoyer, and Veselin Stoyanov. 2019{\natexlab{c}}.
\newblock \href {https://arxiv.org/abs/1907.11692} {{RoBERTa: A} robustly
  optimized {BERT} pretraining approach}.
\newblock \emph{CoRR}, abs/1907.11692.

\bibitem[{Mickus et~al.(2020)Mickus, Paperno, Constant, and van
  Deemter}]{Mickus:2020arxiv}
Timothee Mickus, Denis Paperno, Mathieu Constant, and Kees van Deemter. 2020.
\newblock \href {https://scholarworks.umass.edu/scil/vol3/iss1/34} {What do you
  mean, {BERT}? {A}ssessing {BERT} as a distributional semantics model}.
\newblock \emph{Proceedings of the Society for Computation in Linguistics},
  3(34).

\bibitem[{Mikolov et~al.(2013{\natexlab{a}})Mikolov, Le, and
  Sutskever}]{Mikolov:2013arxiv}
Tomas Mikolov, Quoc~V. Le, and Ilya Sutskever. 2013{\natexlab{a}}.
\newblock \href {https://arxiv.org/pdf/1309.4168.pdf} {Exploiting similarities
  among languages for machine translation}.
\newblock \emph{arXiv preprint, CoRR}, abs/1309.4168.

\bibitem[{Mikolov et~al.(2013{\natexlab{b}})Mikolov, Sutskever, Chen, Corrado,
  and Dean}]{Mikolov:2013nips}
Tomas Mikolov, Ilya Sutskever, Kai Chen, Gregory~S. Corrado, and Jeffrey Dean.
  2013{\natexlab{b}}.
\newblock \href {https://arxiv.org/abs/1310.4546} {Distributed representations
  of words and phrases and their compositionality}.
\newblock In \emph{Proceedings of NeurIPS}, pages 3111--3119.

\bibitem[{Pennington et~al.(2014)Pennington, Socher, and
  Manning}]{Pennington:2014emnlp}
Jeffrey Pennington, Richard Socher, and Christopher Manning. 2014.
\newblock \href {http://www.aclweb.org/anthology/D14-1162} {{Glove: G}lobal
  vectors for word representation}.
\newblock In \emph{Proceedings of EMNLP}, pages 1532--1543.

\bibitem[{Peters et~al.(2018)Peters, Neumann, Iyyer, Gardner, Clark, Lee, and
  Zettlemoyer}]{peters2018deep}
Matthew Peters, Mark Neumann, Mohit Iyyer, Matt Gardner, Christopher Clark,
  Kenton Lee, and Luke Zettlemoyer. 2018.
\newblock \href {https://www.aclweb.org/anthology/N18-1202.pdf} {Deep
  contextualized word representations}.
\newblock In \emph{Proceedings of NAACL-HLT}, pages 2227--2237.

\bibitem[{Pfeiffer et~al.(2020)Pfeiffer, Vuli\'{c}, Gurevych, and
  Ruder}]{Pfeiffer:2020arxiv}
Jonas Pfeiffer, Ivan Vuli\'{c}, Iryna Gurevych, and Sebastian Ruder. 2020.
\newblock \href {https://arxiv.org/abs/2005.00052} {{MAD-X:} {A}n adapter-based
  framework for multi-task cross-lingual transfer}.
\newblock In \emph{Proceedings of EMNLP}.

\bibitem[{Pimentel et~al.(2020)Pimentel, Valvoda, Maudslay, Zmigrod, Williams,
  and Cotterell}]{pimentel2020informationtheoretic}
Tiago Pimentel, Josef Valvoda, Rowan~Hall Maudslay, Ran Zmigrod, Adina
  Williams, and Ryan Cotterell. 2020.
\newblock \href {https://arxiv.org/pdf/2004.03061.pdf} {Information-theoretic
  probing for linguistic structure}.
\newblock In \emph{Proceedings of ACL}, pages 4609--4622.

\bibitem[{Ponti et~al.(2020)Ponti, Glava\v{s}, Majewska, Liu, Vuli\'{c}, and
  Korhonen}]{Ponti:2020xcopa}
Edoardo~Maria Ponti, Goran Glava\v{s}, Olga Majewska, Qianchu Liu, Ivan
  Vuli\'{c}, and Anna Korhonen. 2020.
\newblock \href {https://arxiv.org/abs/2005.00333} {{XCOPA:} {A} multilingual
  dataset for causal commonsense reasoning}.
\newblock In \emph{Proceedings of EMNLP}.

\bibitem[{Qiao et~al.(2019)Qiao, Xiong, Liu, and Liu}]{Qiao:2019arxiv}
Yifan Qiao, Chenyan Xiong, Zheng{-}Hao Liu, and Zhiyuan Liu. 2019.
\newblock \href {http://arxiv.org/abs/1904.07531} {Understanding the behaviors
  of {BERT} in ranking}.
\newblock \emph{CoRR}, abs/1904.07531.

\bibitem[{Qiu et~al.(2020)Qiu, Sun, Xu, Shao, Dai, and Huang}]{Qiu:2020arxiv}
Xipeng Qiu, Tianxiang Sun, Yige Xu, Yunfan Shao, Ning Dai, and Xuanjing Huang.
  2020.
\newblock \href {https://arxiv.org/abs/2003.08271} {Pre-trained models for
  natural language processing: {A} survey}.
\newblock \emph{CoRR}, abs/2003.08271.

\bibitem[{Raffel et~al.(2019)Raffel, Shazeer, Roberts, Lee, Narang, Matena,
  Zhou, Li, and Liu}]{Raffel:2019:arxiv}
Colin Raffel, Noam Shazeer, Adam Roberts, Katherine Lee, Sharan Narang, Michael
  Matena, Yanqi Zhou, Wei Li, and Peter~J. Liu. 2019.
\newblock \href {http://arxiv.org/abs/1910.10683} {Exploring the limits of
  transfer learning with a unified text-to-text transformer}.
\newblock \emph{CoRR}, abs/1910.10683.

\bibitem[{Reif et~al.(2019)Reif, Yuan, Wattenberg, Viegas, Coenen, Pearce, and
  Kim}]{reif2019visualizing}
Emily Reif, Ann Yuan, Martin Wattenberg, Fernanda~B. Viegas, Andy Coenen, Adam
  Pearce, and Been Kim. 2019.
\newblock \href
  {http://papers.nips.cc/paper/9065-visualizing-and-measuring-the-geometry-of-bert.pdf}
  {Visualizing and measuring the geometry of {BERT}}.
\newblock In \emph{Proceedings of NeurIPS}, pages 8594--8603.

\bibitem[{Rogers et~al.(2020)Rogers, Kovaleva, and
  Rumshisky}]{Rogers:2020arxiv}
Anna Rogers, Olga Kovaleva, and Anna Rumshisky. 2020.
\newblock \href {https://arxiv.org/abs/2002.12327} {A primer in {BERTology:}
  what we know about how {BERT} works}.
\newblock \emph{Transactions of the ACL}.

\bibitem[{Ruder et~al.(2019)Ruder, Vuli\'{c}, and S{\o}gaard}]{Ruder:2019jair}
Sebastian Ruder, Ivan Vuli\'{c}, and Anders S{\o}gaard. 2019.
\newblock \href {https://doi.org/10.1613/jair.1.11640} {A survey of
  cross-lingual embedding models}.
\newblock \emph{Journal of Artificial Intelligence Research}, 65:569--631.

\bibitem[{Singh et~al.(2019)Singh, McCann, Socher, and Xiong}]{Singh:2019ws}
Jasdeep Singh, Bryan McCann, Richard Socher, and Caiming Xiong. 2019.
\newblock \href {https://www.aclweb.org/anthology/D19-6106} {{BERT} is not an
  interlingua and the bias of tokenization}.
\newblock In \emph{Proceedings of the 2nd Workshop on Deep Learning Approaches
  for Low-Resource NLP (DeepLo 2019)}, pages 47--55.

\bibitem[{Smith et~al.(2017)Smith, Turban, Hamblin, and
  Hammerla}]{Smith:2017iclr}
Samuel~L. Smith, David~H.P. Turban, Steven Hamblin, and Nils~Y. Hammerla. 2017.
\newblock \href {http://arxiv.org/abs/1702.03859} {Offline bilingual word
  vectors, orthogonal transformations and the inverted softmax}.
\newblock In \emph{Proceedings of ICLR (Conference Track)}.

\bibitem[{S{\o}gaard et~al.(2018)S{\o}gaard, Ruder, and
  Vuli{\'c}}]{Sogaard:2018acl}
Anders S{\o}gaard, Sebastian Ruder, and Ivan Vuli{\'c}. 2018.
\newblock \href {https://www.aclweb.org/anthology/P18-1072} {On the limitations
  of unsupervised bilingual dictionary induction}.
\newblock In \emph{Proceedings of ACL}, pages 778--788.

\bibitem[{Tenney et~al.(2019)Tenney, Das, and Pavlick}]{Tenney:2019acl}
Ian Tenney, Dipanjan Das, and Ellie Pavlick. 2019.
\newblock \href {https://www.aclweb.org/anthology/P19-1452} {{BERT} rediscovers
  the classical {NLP} pipeline}.
\newblock In \emph{Proceedings of ACL}, pages 4593--4601.

\bibitem[{Tiedemann(2009)}]{Tiedemann:2009opus}
J\"org Tiedemann. 2009.
\newblock \href {http://stp.lingfil.uu.se/~joerg/published/ranlp-V.pdf} {News
  from {OPUS} - {A} collection of multilingual parallel corpora with tools and
  interfaces}.
\newblock In \emph{Proceedings of RANLP}, pages 237--248.

\bibitem[{Tsai et~al.(2019)Tsai, Riesa, Johnson, Arivazhagan, Li, and
  Archer}]{Tsai:2019emnlp}
Henry Tsai, Jason Riesa, Melvin Johnson, Naveen Arivazhagan, Xin Li, and Amelia
  Archer. 2019.
\newblock \href {https://www.aclweb.org/anthology/D19-1374} {Small and
  practical {BERT} models for sequence labeling}.
\newblock In \emph{Proceedings EMNLP-IJCNLP}, pages 3632--3636.

\bibitem[{Vaswani et~al.(2017)Vaswani, Shazeer, Parmar, Uszkoreit, Jones,
  Gomez, Kaiser, and Polosukhin}]{Vaswani:2017nips}
Ashish Vaswani, Noam Shazeer, Niki Parmar, Jakob Uszkoreit, Llion Jones,
  Aidan~N. Gomez, Lukasz Kaiser, and Illia Polosukhin. 2017.
\newblock \href {http://papers.nips.cc/paper/7181-attention-is-all-you-need}
  {Attention is all you need}.
\newblock In \emph{Proceedings of NeurIPS}, pages 6000--6010.

\bibitem[{Voita and Titov(2020)}]{voita2020informationtheoretic}
Elena Voita and Ivan Titov. 2020.
\newblock \href {https://arxiv.org/abs/2003.12298} {Information-theoretic
  probing with minimum description length}.
\newblock In \emph{Proceedings of EMNLP}.

\bibitem[{Vuli\'c et~al.(2020)Vuli\'c, Baker, Ponti, Petti, Leviant, Wing,
  Majewska, Bar, Malone, Poibeau, Reichart, and Korhonen}]{Vulic:2020msimlex}
Ivan Vuli\'c, Simon Baker, Edoardo~Maria Ponti, Ulla Petti, Ira Leviant, Kelly
  Wing, Olga Majewska, Eden Bar, Matt Malone, Thierry Poibeau, Roi Reichart,
  and Anna Korhonen. 2020.
\newblock \href {https://arxiv.org/abs/2003.04866} {Multi-{S}imlex: {A}
  large-scale evaluation of multilingual and cross-lingual lexical semantic
  similarity}.
\newblock \emph{Computational Linguistics}.

\bibitem[{Wang et~al.(2020)Wang, Tang, Duan, Wei, Huang, Ji, Cao, Jiang, and
  Zhou}]{Wang:2020kadapter}
Ruize Wang, Duyu Tang, Nan Duan, Zhongyu Wei, Xuanjing Huang, Jianshu Ji,
  Guihong Cao, Daxin Jiang, and Ming Zhou. 2020.
\newblock \href {https://arxiv.org/abs/2002.01808} {{K-Adapter: I}nfusing
  knowledge into pre-trained models with adapters}.
\newblock \emph{CoRR}, abs/2002.01808.

\bibitem[{Wang et~al.(2019)Wang, Che, Guo, Liu, and Liu}]{Wang:2019emnlp}
Yuxuan Wang, Wanxiang Che, Jiang Guo, Yijia Liu, and Ting Liu. 2019.
\newblock \href {https://www.aclweb.org/anthology/D19-1575} {Cross-lingual
  {BERT} transformation for zero-shot dependency parsing}.
\newblock In \emph{Proceedings of EMNLP-IJCNLP}, pages 5721--5727.

\bibitem[{Wolf et~al.(2019)Wolf, Debut, Sanh, Chaumond, Delangue, Moi, Cistac,
  Rault, Louf, Funtowicz, and Brew}]{Wolf:2019hf}
Thomas Wolf, Lysandre Debut, Victor Sanh, Julien Chaumond, Clement Delangue,
  Anthony Moi, Pierric Cistac, Tim Rault, R'emi Louf, Morgan Funtowicz, and
  Jamie Brew. 2019.
\newblock \href {https://arxiv.org/pdf/1910.03771.pdf} {{HuggingFace's
  Transformers: S}tate-of-the-art natural language processing}.
\newblock \emph{ArXiv}, abs/1910.03771.

\bibitem[{Wu et~al.(2020)Wu, Conneau, Li, Zettlemoyer, and
  Stoyanov}]{Wu:2019arxiv}
Shijie Wu, Alexis Conneau, Haoran Li, Luke Zettlemoyer, and Veselin Stoyanov.
  2020.
\newblock \href {http://arxiv.org/abs/1911.01464} {Emerging cross-lingual
  structure in pretrained language models}.
\newblock In \emph{Proceedings of ACL}, pages 6022--6034.

\bibitem[{Wu and Dredze(2019)}]{Wu:2019emnlp}
Shijie Wu and Mark Dredze. 2019.
\newblock \href {https://www.aclweb.org/anthology/D19-1077} {Beto, bentz,
  becas: {T}he surprising cross-lingual effectiveness of {BERT}}.
\newblock In \emph{Proceedings of EMNLP}, pages 833--844.

\bibitem[{Yin and Sch{\"{u}}tze(2016)}]{Yin:2016acl}
Wenpeng Yin and Hinrich Sch{\"{u}}tze. 2016.
\newblock \href {https://doi.org/10.18653/v1/p16-1128} {Learning word
  meta-embeddings}.
\newblock In \emph{Proceedings of ACL}, pages 1351--1360.

\bibitem[{Ziemski et~al.(2016)Ziemski, Junczys{-}Dowmunt, and
  Pouliquen}]{Ziemski:2016lrec}
Michal Ziemski, Marcin Junczys{-}Dowmunt, and Bruno Pouliquen. 2016.
\newblock \href
  {http://www.lrec-conf.org/proceedings/lrec2016/summaries/1195.html} {{The
  United Nations Parallel Corpus v1.0}}.
\newblock In \emph{Proceedings of LREC}.

\end{thebibliography}
\bibliographystyle{acl_natbib}

\clearpage
\appendix
\section{Appendix}

URLs to the models and external corpora used in our study are provided in Table~\ref{tab:models} and Table~\ref{tab:excorpora}, respectively. URLs to the evaluation data and task architectures for each evaluation task are provided in Table~\ref{tab:evtasks}. We also report additional and more detailed sets of results across different tasks, word embedding extraction configurations/variants, and language pairs:

\begin{itemize}
    \item In Table~\ref{tab:full-bli1} and Table~\ref{tab:full-bli2}, we provide full BLI results per language pair. All scores are Mean Reciprocal Rank (MRR) scores (in the standard scoring interval, 0.0--1.0).
    \item In Table~\ref{tab:full-clir}, we provide full CLIR results per language pair. All scores are Mean Average Precision (MAP) scores (in the standard scoring interval, 0.0--1.0).
    \item In Table~\ref{tab:full-stm}, we provide full relation prediction (RELP) results for \en and \de. All scores are micro-averaged $F_1$ scores over 5 runs of the relation predictor \cite{Glavas:2018naacl}. We also report standard deviation for each configuration.
\end{itemize}

\begin{figure*}[!t]
    \centering
    \begin{subfigure}[!ht]{0.488\linewidth}
        \centering
        \includegraphics[width=0.98\linewidth]{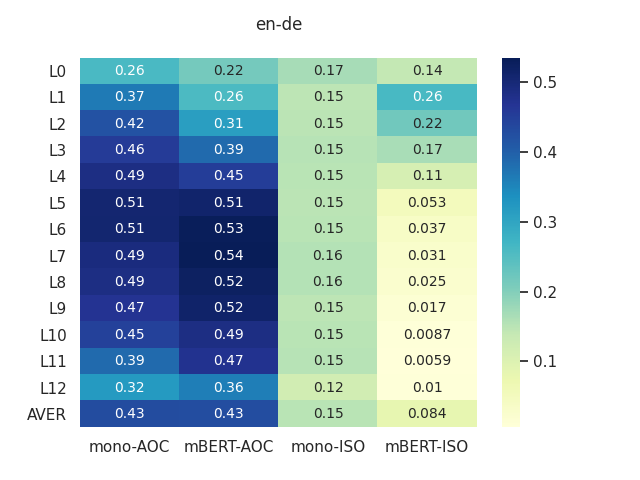}
        \caption{\en--\de: Word translation pairs}
        \label{fig:en-de}
    \end{subfigure}
    \begin{subfigure}[!ht]{0.488\textwidth}
        \centering
        \includegraphics[width=0.98\linewidth]{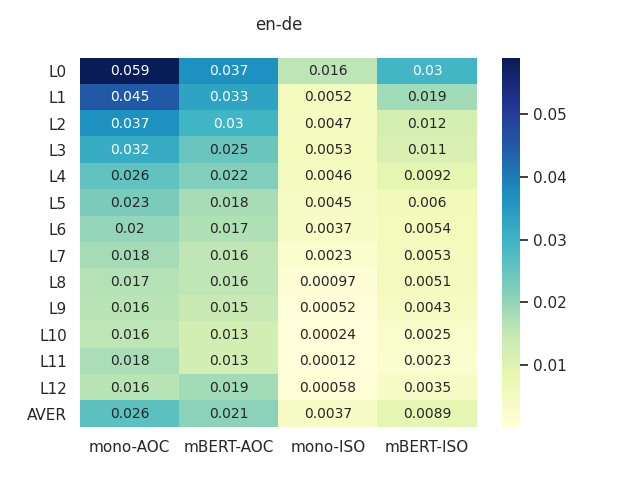}
        \caption{\en--\de: Random word pairs}
        \label{fig:en-de-random}
    \end{subfigure}
    \vspace{-2mm}
    \caption{CKA similarity scores of type-level word representations extracted from each layer (using different extraction configurations, see Table~\ref{tab:config}) for a set of \textbf{(a)} 7K \en--\de translation pairs from the BLI dictionaries of \newcite{Glavas:2019acl}; \textbf{(b)} 7K random \en--\de pairs.}
    \vspace{-1.5mm}
\label{fig:appendix-cka1}
\end{figure*}

\begin{figure*}[!t]
    \centering
    \begin{subfigure}[!ht]{0.488\linewidth}
        \centering
        \includegraphics[width=0.98\linewidth]{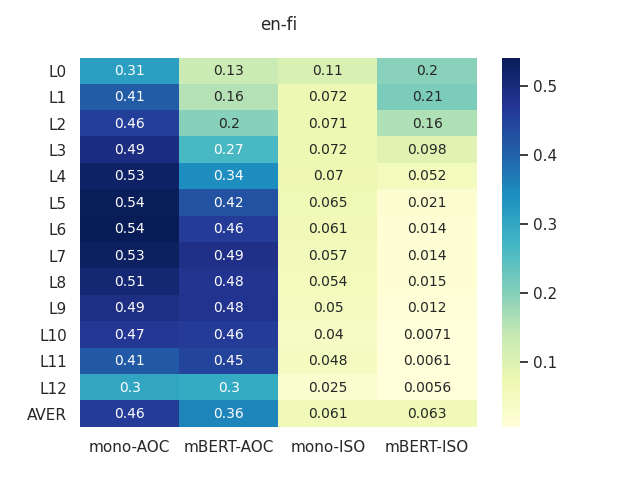}
        \caption{\en--\fin: Word translation pairs}
        \label{fig:en-fi}
    \end{subfigure}
    \begin{subfigure}[!ht]{0.488\textwidth}
        \centering
        \includegraphics[width=0.98\linewidth]{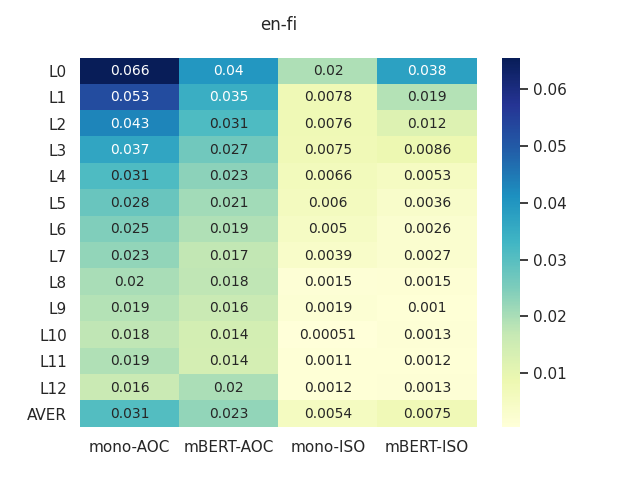}
        \caption{\en--\fin: Random word pairs}
        \label{fig:en-fi-random}
    \end{subfigure}
    \vspace{-2mm}
    \caption{CKA similarity scores of type-level word representations extracted from each layer (using different extraction configurations, see Table~\ref{tab:config}) for a set of \textbf{(a)} 7K \en--\fin translation pairs from the BLI dictionaries of \newcite{Glavas:2019acl}; \textbf{(b)} 7K random \en--\fin pairs.}
    \vspace{-1.5mm}
\label{fig:appendix-cka3}
\end{figure*}

\begin{figure*}[!t]
    \centering
    \begin{subfigure}[!ht]{0.488\linewidth}
        \centering
        \includegraphics[width=0.98\linewidth]{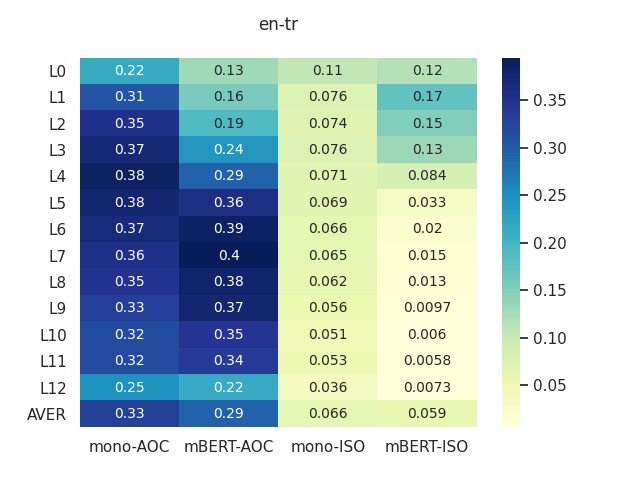}
        \caption{\en--\tr: Word translation pairs}
        \label{fig:en-tr}
    \end{subfigure}
    \begin{subfigure}[!ht]{0.488\textwidth}
        \centering
        \includegraphics[width=0.98\linewidth]{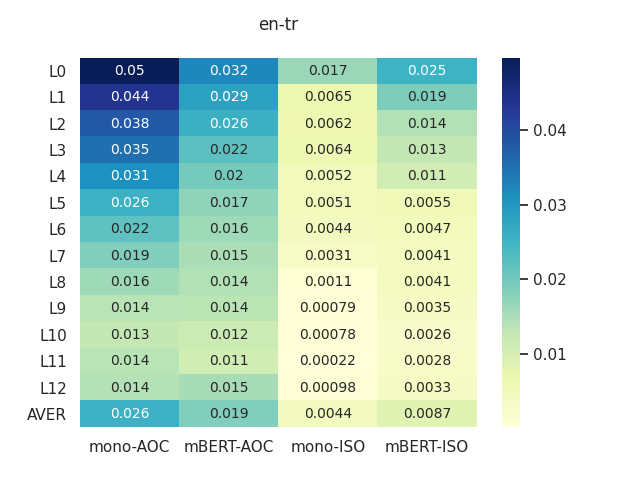}
        \caption{\en--\tr: Random word pairs}
        \label{fig:en-tr-random}
    \end{subfigure}
    \vspace{-2mm}
    \caption{CKA similarity scores of type-level word representations extracted from each layer (using different extraction configurations, see Table~\ref{tab:config}) for a set of \textbf{(a)} 7K \en--\tr translation pairs from the BLI dictionaries of \newcite{Glavas:2019acl}; \textbf{(b)} 7K random \en--\tr pairs.}
    \vspace{-1.5mm}
\label{fig:appendix-cka4}
\end{figure*}

Finally, in Figures~\ref{fig:appendix-cka1}-\ref{fig:appendix-cka4}, we also provide heatmaps denoting bilingual layer correspondence, computed via linear CKA similarity \cite{Kornblith:2019icml}, for several \textsc{en}--$L_t$ language pairs (see \S\ref{ss:individual}), which are not provided in the main paper 

\onecolumn
\begin{table*}[!t]
\def\arraystretch{0.99}
\centering
{\footnotesize
\begin{adjustbox}{max width=\linewidth}
\begin{tabularx}{\textwidth}{l X}
\toprule
{\bf Language} & {\bf URL} \\
\cmidrule(lr){2-2}
{\en} & {\url{https://huggingface.co/bert-base-uncased}} \\
{\de} & {\url{https://huggingface.co/bert-base-german-dbmdz-uncased}} \\
{\ru} & {\url{https://huggingface.co/DeepPavlov/rubert-base-cased}} \\
{\fin} & {\url{https://huggingface.co/TurkuNLP/bert-base-finnish-uncased-v1}} \\
{\zh} & {\url{https://huggingface.co/bert-base-chinese}} \\
{\tr} & {\url{https://huggingface.co/dbmdz/bert-base-turkish-uncased}} \\
{Multilingual} & {\url{https://huggingface.co/bert-base-multilingual-uncased}} \\
\cmidrule(lr){2-2}
\multirow{2}{*}{\textsc{it}} & {\url{https://huggingface.co/dbmdz/bert-base-italian-uncased}} \\
& {\url{https://huggingface.co/dbmdz/bert-base-italian-xxl-uncased}} \\
\bottomrule
\end{tabularx}
\end{adjustbox}
}
\vspace{-1.5mm}
\caption{URLs of the models used in our study. The first part of the table refers to the models used in the main experiments throughout the paper, while the second part refers to the models used in side experiments.}
\label{tab:models}
\end{table*}

\begin{table*}[!t]
\def\arraystretch{0.99}
\centering
{\footnotesize
\begin{adjustbox}{max width=\linewidth}
\begin{tabularx}{\textwidth}{l X}
\toprule
{\bf Language} & {\bf URL} \\
\cmidrule(lr){2-2}
{\en} & {\url{http://opus.nlpl.eu/download.php?f=Europarl/v8/moses/de-en.txt.zip}} \\
{\de} & {\url{http://opus.nlpl.eu/download.php?f=Europarl/v8/moses/de-en.txt.zip}} \\
{\ru} & {\url{http://opus.nlpl.eu/download.php?f=UNPC/v1.0/moses/en-ru.txt.zip}} \\
{\fin} & {\url{http://opus.nlpl.eu/download.php?f=Europarl/v8/moses/en-fi.txt.zip}} \\
{\zh} & {\url{http://opus.nlpl.eu/download.php?f=UNPC/v1.0/moses/en-zh.txt.zip}} \\
{\tr} & {\url{http://data.statmt.org/wmt18/translation-task/news.2017.tr.shuffled.deduped.gz}} \\
 \textsc{it} & {\url{http://opus.nlpl.eu/download.php?f=Europarl/v8/moses/en-it.txt.zip}} \\
\bottomrule
\end{tabularx}
\end{adjustbox}
}
\vspace{-1.5mm}
\caption{Links to the external corpora used in the study. We randomly sample 1M sentences of maximum sequence length 512 from the corresponding corpora.}
\label{tab:excorpora}
\end{table*}
\begin{table*}[!t]
\def\arraystretch{0.99}
\centering
{\footnotesize
\begin{adjustbox}{max width=\linewidth}
\begin{tabularx}{\textwidth}{l X X}
\toprule
{\bf Task} & {\bf Evaluation Data and/or Model} & {\bf Link} \\
\cmidrule(lr){2-2} \cmidrule(lr){3-3}
{LSIM} & {Multi-SimLex} & {Data: \url{multisimlex.com/}} \\
\cmidrule(lr){2-2} \cmidrule(lr){3-3}
{WA} & {BATS} & {Data: \url{vecto.space/projects/BATS/}} \\
\cmidrule(lr){2-2} \cmidrule(lr){3-3}
\multirow{2}{*}{BLI} & {Data: Dictionaries from \newcite{Glavas:2019acl}} & {Data: \url{github.com/codogogo/xling-eval/tree/master/bli_datasets}} \\
& {Model: VecMap}  & {Model: \url{github.com/artetxem/vecmap}} \\
\cmidrule(lr){2-2} \cmidrule(lr){3-3}
\multirow{2}{*}{CLIR} & {Data: CLEF 2003} & {Data: \url{catalog.elra.info/en-us/repository/browse/ELRA-E0008/}} \\
& {Model: Agg-IDF from \newcite{Litschko:2019sigir}}  & {Model: \url{github.com/rlitschk/UnsupCLIR}} \\
\cmidrule(lr){2-2} \cmidrule(lr){3-3}
\multirow{2}{*}{RELP} & {Data: WordNet-based RELP data} & {Data: \url{github.com/codogogo/stm/tree/master/data/wn-ls}} \\
& {Model: Specialization Tensor Model}  & {Model: \url{github.com/codogogo/stm}} \\
\bottomrule
\end{tabularx}
\end{adjustbox}
}
\vspace{-1.5mm}
\caption{Links to evaluation data and models.}
\label{tab:evtasks}
\end{table*}


\begin{table*}[t]
\def\arraystretch{0.93}
\centering
{\footnotesize
\begin{adjustbox}{max width=\linewidth}
\begin{tabularx}{\textwidth}{l YYYYYY Y}
\toprule
{\textbf{Configuration}} & {\en--\de} & {\en--\tr} & {\en--\fin} & {\en--\ru} & {\de--\tr} & {\de--\fin} & {\de--\ru} \\
\cmidrule(lr){2-8}
\textsc{fasttext.wiki} & {0.610} & {0.433} & {0.488} & {0.522} & {0.358} & {0.435} & {0.469} \\
\cmidrule(lr){2-8}
\rowcolor{Gray}
{\monobert.\iso.\nospec} & {} & {} & {} & {} & {} & {} & {} \\
{\textsc{avg(}L$\leq$2)} & 0.390 & 0.332 & 0.392 & 0.409 & 0.237 & 0.269 & 0.291 \\
{\textsc{avg(}L$\leq$4)} & 0.430 & 0.367 & 0.438 & 0.447 & 0.269 & 0.311 & 0.338 \\
{\textsc{avg(}L$\leq$6)} & 0.461 & 0.386 & 0.476 & 0.472 & 0.299 & 0.359 & 0.387 \\
{\textsc{avg(}L$\leq$8)} & 0.472 & 0.390 & 0.486 & 0.487 & 0.315 & 0.387 & 0.407 \\
{\textsc{avg(}L$\leq$10)} & 0.461 & 0.386 & 0.483 & 0.488 & 0.321 & 0.395 & 0.416 \\
{\textsc{avg(}L$\leq$12)} & 0.446 & 0.379 & 0.471 & 0.473 & 0.323 & 0.395 & 0.412 \\
\rowcolor{Gray}
{\monobert.\aoc-\tens.\nospec} & {} & {} & {} & {} & {} & {} & {} \\
{\textsc{avg(}L$\leq$2)} & 0.399 & 0.342 & 0.386 & 0.403 & 0.242 & 0.269 & 0.292 \\
{\textsc{avg(}L$\leq$4)} & 0.457 & 0.379 & 0.448 & 0.433 & 0.283 & 0.322 & 0.343 \\
{\textsc{avg(}L$\leq$6)} & 0.503 & 0.399 & 0.480 & 0.458 & 0.315 & 0.369 & 0.380 \\
{\textsc{avg(}L$\leq$8)} & 0.527 & 0.414 & 0.499 & 0.461 & 0.332 & 0.394 & 0.391 \\
{\textsc{avg(}L$\leq$10)} & 0.534 & 0.415 & 0.498 & 0.459 & 0.337 & 0.401 & 0.394 \\
{\textsc{avg(}L$\leq$12)} & 0.534 & 0.416 & 0.492 & 0.453 & 0.337 & 0.401 & 0.376 \\
\rowcolor{Gray}
{\monobert.\aoc-\huns.\nospec} & {} & {} & {} & {} & {} & {} & {} \\
{\textsc{avg(}L$\leq$2)} & 0.401 & 0.343 & 0.391 & 0.398 & 0.239 & 0.269 & 0.293 \\
{\textsc{avg(}L$\leq$4)} & 0.459 & 0.381 & 0.449 & 0.437 & 0.288 & 0.325 & 0.343 \\
{\textsc{avg(}L$\leq$6)}& 0.504 & 0.403 & 0.484 & 0.459 & 0.318 & 0.373 & 0.382 \\
{\textsc{avg(}L$\leq$8)} & 0.532 & 0.418 & 0.503 & 0.462 & 0.334 & 0.394 & 0.389 \\
{\textsc{avg(}L$\leq$10)} & 0.540 & 0.422 & {0.504} & 0.459 & 0.338 & 0.402 & 0.393 \\
{\textsc{avg(}L$\leq$12)} & 0.542 & 0.426 & 0.500 & 0.454 & 0.343 & 0.401 & 0.378 \\
\rowcolor{Gray}
{\monobert.\iso.\all} & {} & {} & {} & {} & {} & {} & {} \\
{\textsc{avg(}L$\leq$2)} & 0.352 & 0.289 & 0.351 & 0.374 & 0.230 & 0.265 & 0.283 \\
{\textsc{avg(}L$\leq$4)} & 0.375 & 0.317 & 0.391 & 0.393 & 0.264 & 0.302 & 0.331 \\
{\textsc{avg(}L$\leq$6)} & 0.386 & 0.330 & 0.406 & 0.407 & 0.289 & 0.350 & 0.376 \\
{\textsc{avg(}L$\leq$8)} & 0.372 & 0.327 & 0.409 & 0.413 & 0.291 & 0.370 & 0.392 \\
{\textsc{avg(}L$\leq$10)} & 0.352 & 0.320 & 0.396 & 0.402 & 0.290 & 0.370 & 0.383 \\
{\textsc{avg(}L$\leq$12)} & 0.313 & 0.310 & 0.373 & 0.394 & 0.283 & 0.358 & 0.371 \\
\rowcolor{Gray}
{\monobert.\iso.\cls} & {} & {} & {} & {} & {} & {} & {} \\
{\textsc{avg(}L$\leq$2)} & 0.367 & 0.306 & 0.368 & 0.386 & 0.236 & 0.272 & 0.285 \\
{\textsc{avg(}L$\leq$4)} & 0.394 & 0.339 & 0.408 & 0.410 & 0.267 & 0.307 & 0.331 \\
{\textsc{avg(}L$\leq$6)} & 0.406 & 0.344 & 0.428 & 0.425 & 0.294 & 0.353 & 0.381 \\
{\textsc{avg(}L$\leq$8)} & 0.393 & 0.344 & 0.430 & 0.431 & 0.306 & 0.369 & 0.400 \\
{\textsc{avg(}L$\leq$10)} & 0.371 & 0.336 & 0.421 & 0.421 & 0.303 & 0.382 & 0.395 \\
{\textsc{avg(}L$\leq$12)}  & 0.331 & 0.329 & 0.403 & 0.409 & 0.302 & 0.375 & 0.387 \\
\rowcolor{Gray}
{\mbert.\iso.\nospec} & {} & {} & {} & {} & {} & {} & {} \\
{\textsc{avg(}L$\leq$2)} & 0.293 & 0.176 & 0.176 & 0.147 & 0.216 & 0.203 & 0.160 \\
{\textsc{avg(}L$\leq$4)} & 0.304 & 0.184 & 0.190 & 0.164 & 0.219 & 0.214 & 0.178 \\
{\textsc{avg(}L$\leq$6)} & 0.315 & 0.189 & 0.203 & 0.198 & 0.223 & 0.225 & 0.198 \\
{\textsc{avg(}L$\leq$8)} & 0.325 & 0.193 & 0.209 & 0.228 & 0.224 & 0.235 & 0.217 \\
{\textsc{avg(}L$\leq$10)} & 0.330 & 0.194 & 0.210 & 0.243 & 0.220 & 0.234 & 0.226 \\
{\textsc{avg(}L$\leq$12)} & 0.333 & 0.193 & 0.206 & 0.248 & 0.219 & 0.231 & 0.227 \\
\rowcolor{Gray}
{\mbert.\aoc-\tens.\nospec} & {} & {} & {} & {} & {} & {} & {} \\
{\textsc{avg(}L$\leq$2)} & 0.309 & 0.171 & 0.172 & 0.146 & 0.208 & 0.200 & 0.156 \\
{\textsc{avg(}L$\leq$4)} & 0.350 & 0.186 & 0.189 & 0.186 & 0.224 & 0.214 & 0.191 \\
{\textsc{avg(}L$\leq$6)}  & 0.389 & 0.219 & 0.215 & 0.240 & 0.241 & 0.243 & 0.225 \\
{\textsc{avg(}L$\leq$8)} & 0.432 & 0.246 & 0.251 & 0.287 & 0.255 & 0.263 & 0.254 \\
{\textsc{avg(}L$\leq$10)} & 0.448 & 0.258 & 0.264 & 0.306 & 0.260 & 0.282 & 0.272 \\
{\textsc{avg(}L$\leq$12)} & 0.456 & 0.267 & 0.272 & 0.316 & 0.260 & 0.292 & 0.284 \\
\rowcolor{Gray}
{\mbert.\iso.\all} & {} & {} & {} & {} & {} & {} & {} \\
{\textsc{avg(}L$\leq$2)} & 0.292 & 0.173 & 0.175 & 0.143 & 0.209 & 0.203 & 0.154 \\
{\textsc{avg(}L$\leq$4)} & 0.301 & 0.176 & 0.188 & 0.155 & 0.211 & 0.213 & 0.171 \\
{\textsc{avg(}L$\leq$6)} & 0.307 & 0.181 & 0.198 & 0.186 & 0.216 & 0.221 & 0.193 \\
{\textsc{avg(}L$\leq$8)} & 0.315 & 0.184 & 0.202 & 0.207 & 0.213 & 0.228 & 0.208 \\
{\textsc{avg(}L$\leq$10)} & 0.318 & 0.182 & 0.197 & 0.216 & 0.208 & 0.226 & 0.215 \\
{\textsc{avg(}L$\leq$12)} & 0.319 & 0.181 & 0.189 & 0.220 & 0.209 & 0.220 & 0.213 \\
\midrule
\rowcolor{Gray}
{\monobert.\iso.\nospec (\textsc{reverse})} & {} & {} & {} & {} & {} & {} & {} \\
{\textsc{avg(}L$\geq$12)} & {0.104} & {--} & {0.054} & {--} & {--} & {0.077} & {--} \\
{\textsc{avg(}L$\geq$10)} & {0.119} & {--} & {0.061} & {--} & {--} & {0.063} & {--} \\
{\textsc{avg(}L$\geq$8)} & {0.144} & {--} & {0.108} & {--} & {--} & {0.095} & {--} \\
{\textsc{avg(}L$\geq$6)} & {0.230} & {--} & {0.223} & {--} & {--} & {0.238} & {--} \\
{\textsc{avg(}L$\geq$4)} & {0.308} & {--} & {0.318} & {--} & {--} & {0.335} & {--} \\
{\textsc{avg(}L$\geq$2)} & {0.365} & {--} & {0.385} & {--} & {--} & {0.372} & {--} \\
{\textsc{avg(}L$\geq$0)} & {0.446} & {--} & {0.471} & {--} & {--} & {0.395} & {--} \\
\bottomrule
\end{tabularx}
\end{adjustbox}
}
\vspace{-1.5mm}
\caption{Results in the BLI task across different language pairs and word vector extraction configurations. MRR scores reported. For clarity of presentation, a subset of results is presented in this table, while the rest (and the averages) are presented in Table~\ref{tab:full-bli2}. \textit{\textsc{avg(}L$\leq$n)} means that we average representations over all Transformer layers up to the $n$th layer (included), where $L=0$ refers to the embedding layer, $L=1$ to the bottom layer, and $L=12$ to the final (top) layer. Different configurations are described in \S\ref{s:methodology} and Table~\ref{tab:config}. Additional diagnostic experiments with top-to-bottom layerwise averaging configs (\textsc{reverse}) are run for a subset of languages: \{\en, \de, \fin\}.}
\label{tab:full-bli1}
\end{table*}

\begin{table*}[t]
\def\arraystretch{0.93}
\centering
{\footnotesize
\begin{adjustbox}{max width=\linewidth}
\begin{tabularx}{\textwidth}{l YYY Y}
\toprule
{\textbf{Configuration}} & {\tr--\fin} & {\tr--\ru} & {\fin--\ru} & {{average}} \\
\cmidrule(lr){2-5}
\textsc{fasttext.wiki} & {0.358} & {0.364} & {0.439} & {0.448} \\
\cmidrule(lr){2-5}
\rowcolor{Gray}
{\monobert.\iso.\nospec} & {} & {} & {} & {} \\
{\textsc{avg(}L$\leq$2)} & 0.237 & 0.217 & 0.290 & 0.306 \\
{\textsc{avg(}L$\leq$4)} & 0.279 & 0.261 & 0.337 & 0.348 \\
{\textsc{avg(}L$\leq$6)} & 0.311 & 0.288 & 0.372 & 0.381 \\
{\textsc{avg(}L$\leq$8)} & 0.334 & 0.315 & 0.387 & 0.398 \\
{\textsc{avg(}L$\leq$10)} & 0.347 & 0.317 & 0.392 & 0.401 \\
{\textsc{avg(}L$\leq$12)} & 0.352 & 0.319 & 0.387 & 0.396 \\
\rowcolor{Gray}
{\monobert.\aoc-\tens.\nospec} & {} & {} & {} & {} \\
{\textsc{avg(}L$\leq$2)} & 0.247 & 0.221 & 0.284 & 0.308 \\
{\textsc{avg(}L$\leq$4)} & 0.288 & 0.263 & 0.331 & 0.355 \\
{\textsc{avg(}L$\leq$6)} & 0.319 & 0.294 & 0.366 & 0.388 \\
{\textsc{avg(}L$\leq$8)} & 0.334 & 0.311 & 0.375 & 0.404 \\
{\textsc{avg(}L$\leq$10)} & 0.340 & 0.311 & 0.379 & 0.407 \\
{\textsc{avg(}L$\leq$12)} & 0.344 & 0.310 & 0.360 & 0.402 \\
\rowcolor{Gray}
{\monobert.\aoc-\huns.\nospec} & {} & {} & {} & {} \\
{\textsc{avg(}L$\leq$2)} & 0.244 & 0.220 & 0.285 & 0.308 \\
{\textsc{avg(}L$\leq$4)} & 0.288 & 0.261 & 0.333 & 0.356 \\
{\textsc{avg(}L$\leq$6)} & 0.322 & 0.291 & 0.367 & 0.390 \\
{\textsc{avg(}L$\leq$8)} & 0.338 & 0.309 & 0.376 & 0.406 \\
{\textsc{avg(}L$\leq$10)} & 0.348 & 0.314 & 0.377 & 0.410 \\
{\textsc{avg(}L$\leq$12)} & 0.349 & 0.311 & 0.361 & 0.407 \\
\rowcolor{Gray}
{\monobert.\iso.\all} & {} & {} & {} & {} \\
{\textsc{avg(}L$\leq$2)} & 0.226 & 0.212 & 0.284 & 0.287 \\
{\textsc{avg(}L$\leq$4)} & 0.270 & 0.254 & 0.328 & 0.322 \\
{\textsc{avg(}L$\leq$6)} & 0.302 & 0.274 & 0.358 & 0.348 \\
{\textsc{avg(}L$\leq$8)} & 0.318 & 0.296 & 0.371 & 0.356 \\
{\textsc{avg(}L$\leq$10)} & 0.328 & 0.303 & 0.373 & 0.352 \\
{\textsc{avg(}L$\leq$12)} & 0.328 & 0.306 & 0.368 & 0.340 \\
\rowcolor{Gray}
{\monobert.\iso.\cls} & {} & {} & {} & {} \\
{\textsc{avg(}L$\leq$2)} & 0.232 & 0.217 & 0.285 & 0.295 \\
{\textsc{avg(}L$\leq$4)} & 0.274 & 0.257 & 0.331 & 0.332 \\
{\textsc{avg(}L$\leq$6)}& 0.307 & 0.279 & 0.362 & 0.358 \\
{\textsc{avg(}L$\leq$8)} & 0.327 & 0.303 & 0.377 & 0.368 \\
{\textsc{avg(}L$\leq$10)} & 0.334 & 0.314 & 0.383 & 0.366 \\
{\textsc{avg(}L$\leq$12)} & 0.340 & 0.317 & 0.373 & 0.357 \\
\rowcolor{Gray}
{\mbert.\iso.\nospec} & {} & {} & {} & {} \\
{\textsc{avg(}L$\leq$2)} & 0.170 & 0.131 & 0.127 & 0.180 \\
{\textsc{avg(}L$\leq$4)} & 0.180 & 0.135 & 0.138 & 0.191 \\
{\textsc{avg(}L$\leq$6)} & 0.188 & 0.147 & 0.151 & 0.204 \\
{\textsc{avg(}L$\leq$8)} & 0.189 & 0.152 & 0.164 & 0.214 \\
{\textsc{avg(}L$\leq$10)} & 0.188 & 0.153 & 0.165 & 0.216 \\
{\textsc{avg(}L$\leq$12)} & 0.188 & 0.158 & 0.163 & 0.217 \\
\rowcolor{Gray}
{\mbert.\aoc-\tens.\nospec} & {} & {} & {} & {} \\
{\textsc{avg(}L$\leq$2)}& 0.165 & 0.127 & 0.130 & 0.178 \\
{\textsc{avg(}L$\leq$4)} & 0.176 & 0.146 & 0.139 & 0.200 \\
{\textsc{avg(}L$\leq$6)} & 0.192 & 0.174 & 0.162 & 0.230 \\
{\textsc{avg(}L$\leq$8)} & 0.210 & 0.192 & 0.185 & 0.258 \\
{\textsc{avg(}L$\leq$10)} & 0.219 & 0.198 & 0.200 & 0.271 \\
{\textsc{avg(}L$\leq$12)} & 0.223 & 0.198 & 0.206 & 0.277 \\
\rowcolor{Gray}
{\mbert.\iso.\all} & {} & {} & {} & {} \\
{\textsc{avg(}L$\leq$2)} & 0.163 & 0.126 & 0.123 & 0.176 \\
{\textsc{avg(}L$\leq$4)} & 0.175 & 0.128 & 0.133 & 0.185 \\
{\textsc{avg(}L$\leq$6)} & 0.179 & 0.139 & 0.142 & 0.196 \\
{\textsc{avg(}L$\leq$8)} & 0.182 & 0.144 & 0.152 & 0.203 \\
{\textsc{avg(}L$\leq$10)} & 0.178 & 0.141 & 0.153 & 0.203 \\
{\textsc{avg(}L$\leq$12)} & 0.175 & 0.143 & 0.150 & 0.202 \\
\bottomrule
\end{tabularx}
\end{adjustbox}
}
\vspace{-1.5mm}
\caption{Results in the bilingual lexicon induction (BLI) task across different language pairs and word vector extraction configurations: Part II. MAP scores reported. For clarity of presentation, a subset of results is presented in this table, while the rest (also used to calculate the averages) is provided in Table~\ref{tab:full-bli1} in the previous page. \textit{\textsc{avg(}L$\leq$n)} means that we average representations over all Transformer layers up to the $n$th layer (included), where $L=0$ refers to the embedding layer, $L=1$ to the bottom layer, and $L=12$ to the final (top) layer. Different configurations are described in \S\ref{s:methodology} and Table~\ref{tab:config}.}
\label{tab:full-bli2}
\end{table*}

\begin{table*}[t]
\def\arraystretch{0.93}
\centering
{\footnotesize
\begin{adjustbox}{max width=\linewidth}
\begin{tabularx}{\textwidth}{l YYYYYY Y}
\toprule
{\textbf{Configuration}} & {\en--\de} & {\en--\fin} & {\en--\ru} & {\de--\fin} & {\de--\ru} & {\fin--\ru} & {{average}} \\
\cmidrule(lr){2-8}
\textsc{fasttext.wiki} & {\bf 0.193} & {\bf 0.136} & {0.118} & {\bf 0.221} & {\bf 0.112} & {0.105} & {\bf 0.148} \\
\cmidrule(lr){2-8}
\rowcolor{Gray}
{\monobert.\iso.\nospec} & {} & {} & {} & {} & {} & {} & {} \\
{\textsc{avg(}L$\leq$2)} & 0.059 & 0.075 & 0.106 & 0.126 & 0.086 & 0.123 & 0.096 \\
{\textsc{avg(}L$\leq$4)} & 0.061 & 0.069 & 0.098 & 0.111 & 0.075 & 0.106 & 0.087 \\
{\textsc{avg(}L$\leq$6)} & 0.052 & 0.061 & 0.079 & 0.112 & 0.068 & 0.102 & 0.079 \\
{\textsc{avg(}L$\leq$8)}& 0.042 & 0.048 & 0.075 & 0.112 & 0.063 & 0.105 & 0.074 \\
{\textsc{avg(}L$\leq$10)} & 0.036 & 0.043 & 0.067 & 0.107 & 0.065 & 0.080 & 0.066 \\
{\textsc{avg(}L$\leq$12)} & 0.032 & 0.034 & 0.059 & 0.097 & 0.077 & 0.083 & 0.064 \\
\rowcolor{Gray}
{\monobert.\aoc-\tens.\nospec} & {} & {} & {} & {} & {} & {} & {} \\
{\textsc{avg(}L$\leq$2)} & 0.069 & 0.078 & 0.094 & 0.109 & 0.078 & 0.108 & 0.089 \\
{\textsc{avg(}L$\leq$4)} & 0.076 & 0.105 & 0.119 & 0.112 & 0.098 & 0.117 & 0.104 \\
{\textsc{avg(}L$\leq$6)} & 0.086 & 0.090 & 0.129 & 0.122 & 0.098 & 0.125 & 0.108 \\
{\textsc{avg(}L$\leq$8)} & 0.092 & 0.073 & 0.137 & 0.105 & 0.100 & 0.114 & 0.103 \\
{\textsc{avg(}L$\leq$10)} & 0.095 & 0.073 & 0.147 & 0.102 & 0.102 & 0.135 & 0.109 \\
{\textsc{avg(}L$\leq$12)} & 0.104 & 0.073 & 0.139 & 0.100 & 0.105 & 0.131 & 0.109 \\
\rowcolor{Gray}
{\monobert.\aoc-\huns.\nospec} & {} & {} & {} & {} & {} & {} & {} \\
{\textsc{avg(}L$\leq$2)} & 0.073 & 0.081 & 0.097 & 0.111 & 0.078 & 0.106 & 0.091 \\
{\textsc{avg(}L$\leq$4)}& 0.078 & 0.107 & 0.115 & 0.107 & 0.100 & 0.115 & 0.104 \\
{\textsc{avg(}L$\leq$6)} & 0.087 & 0.087 & 0.127 & 0.132 & 0.103 & {\bf 0.123} & 0.110 \\
{\textsc{avg(}L$\leq$8)} & 0.091 & 0.076 & 0.137 & 0.118 & 0.101 & 0.106 & 0.105 \\
{\textsc{avg(}L$\leq$10)} & 0.099 & 0.074 & {\bf 0.161} & 0.103 & 0.104 & 0.104 & 0.107 \\
{\textsc{avg(}L$\leq$12)} & 0.106 & 0.076 & 0.146 & 0.105 & 0.106 & 0.100 & 0.106 \\
\rowcolor{Gray}
{\monobert.\iso.\all} & {} & {} & {} & {} & {} & {} & {} \\
{\textsc{avg(}L$\leq$2)} & 0.044 & 0.045 & 0.076 & 0.095 & 0.067 & 0.098 & 0.071 \\
{\textsc{avg(}L$\leq$4)} & 0.039 & 0.042 & 0.079 & 0.094 & 0.066 & 0.100 & 0.070 \\
{\textsc{avg(}L$\leq$6)} & 0.024 & 0.034 & 0.069 & 0.089 & 0.066 & 0.094 & 0.063 \\
{\textsc{avg(}L$\leq$8)} & 0.018 & 0.020 & 0.039 & 0.068 & 0.059 & 0.092 & 0.049 \\
{\textsc{avg(}L$\leq$10)} & 0.016 & 0.016 & 0.030 & 0.048 & 0.058 & 0.067 & 0.039 \\
{\textsc{avg(}L$\leq$12)}& 0.014 & 0.013 & 0.033 & 0.034 & 0.064 & 0.061 & 0.036 \\
\rowcolor{Gray}
{\monobert.\iso.\cls} & {} & {} & {} & {} & {} & {} & {} \\
{\textsc{avg(}L$\leq$2)} & 0.050 & 0.057 & 0.086 & 0.106 & 0.071 & 0.108 & 0.080 \\
{\textsc{avg(}L$\leq$4)} & 0.046 & 0.055 & 0.084 & 0.104 & 0.071 & 0.102 & 0.077 \\
{\textsc{avg(}L$\leq$6)} & 0.032 & 0.042 & 0.076 & 0.103 & 0.066 & 0.097 & 0.069 \\
{\textsc{avg(}L$\leq$8)} & 0.025 & 0.028 & 0.046 & 0.086 & 0.059 & 0.101 & 0.057 \\
{\textsc{avg(}L$\leq$10)} & 0.021 & 0.030 & 0.037 & 0.072 & 0.057 & 0.079 & 0.049 \\
{\textsc{avg(}L$\leq$12)} & 0.020 & 0.016 & 0.032 & 0.052 & 0.045 & 0.072 & 0.040 \\
\rowcolor{Gray}
{\mbert.\iso.\nospec} & {} & {} & {} & {} & {} & {} & {} \\
{\textsc{avg(}L$\leq$2)} & 0.110 & 0.009 & 0.045 & 0.057 & 0.020 & 0.013 & 0.042 \\
{\textsc{avg(}L$\leq$4)} & 0.100 & 0.007 & 0.075 & 0.044 & 0.025 & 0.011 & 0.044 \\
{\textsc{avg(}L$\leq$6)}& 0.098 & 0.007 & 0.046 & 0.043 & 0.029 & 0.030 & 0.042 \\
{\textsc{avg(}L$\leq$8)} & 0.088 & 0.008 & 0.052 & 0.043 & 0.032 & 0.031 & 0.042 \\
{\textsc{avg(}L$\leq$10)} & 0.084 & 0.008 & 0.051 & 0.042 & 0.034 & 0.026 & 0.041 \\
{\textsc{avg(}L$\leq$12)} & 0.082 & 0.006 & 0.048 & 0.039 & 0.037 & 0.024 & 0.039 \\
\rowcolor{Gray}
{\mbert.\aoc-\tens.\nospec} & {} & {} & {} & {} & {} & {} & {} \\
{\textsc{avg(}L$\leq$2)} & 0.127 & 0.013 & 0.049 & 0.027 & 0.019 & 0.009 & 0.041 \\
{\textsc{avg(}L$\leq$4)} & 0.123 & 0.018 & 0.055 & 0.032 & 0.029 & 0.008 & 0.044 \\
{\textsc{avg(}L$\leq$6)} & 0.120 & 0.018 & 0.055 & 0.051 & 0.042 & 0.009 & 0.049 \\
{\textsc{avg(}L$\leq$8)} & 0.123 & 0.018 & 0.057 & 0.053 & 0.049 & 0.016 & 0.053 \\
{\textsc{avg(}L$\leq$10)} & 0.127 & 0.019 & 0.062 & 0.050 & 0.051 & 0.018 & 0.054 \\
{\textsc{avg(}L$\leq$12)} & 0.128 & 0.021 & 0.065 & 0.049 & 0.052 & 0.019 & 0.056 \\
\rowcolor{Gray}
{\mbert.\iso.\all} & {} & {} & {} & {} & {} & {} & {} \\
{\textsc{avg(}L$\leq$2)} & 0.072 & 0.005 & 0.032 & 0.014 & 0.016 & 0.004 & 0.024 \\
{\textsc{avg(}L$\leq$4)} & 0.075 & 0.004 & 0.027 & 0.014 & 0.022 & 0.005 & 0.024 \\
{\textsc{avg(}L$\leq$6)} & 0.065 & 0.004 & 0.026 & 0.015 & 0.027 & 0.007 & 0.024 \\
{\textsc{avg(}L$\leq$8)} & 0.054 & 0.004 & 0.035 & 0.015 & 0.032 & 0.008 & 0.025 \\
{\textsc{avg(}L$\leq$10)} & 0.054 & 0.005 & 0.032 & 0.017 & 0.035 & 0.007 & 0.025 \\
{\textsc{avg(}L$\leq$12)} & 0.058 & 0.004 & 0.034 & 0.018 & 0.032 & 0.006 & 0.025 \\
\midrule
\rowcolor{Gray}
{\monobert.\iso.\nospec (\textsc{reverse})} & {} & {} & {} & {} & {} & {} & {} \\
{\textsc{avg(}L$\geq$12)} & {0.005} & {0.012} & {--} & {0.001} & {--} & {--} & {--} \\
{\textsc{avg(}L$\geq$10)} & {0.002} & {0.002} & {--} & {0.001} & {--} & {--} & {--} \\
{\textsc{avg(}L$\geq$8)} & {0.004} & {0.002} & {--} & {0.002} & {--} & {--} & {--} \\
{\textsc{avg(}L$\geq$6)} & {0.014} & {0.006} & {--} & {0.004} & {--} & {--} & {--} \\
{\textsc{avg(}L$\geq$4)} & {0.020} & {0.012} & {--} & {0.016} & {--} & {--} & {--} \\
{\textsc{avg(}L$\geq$2)} & {0.024} & {0.019} & {--} & {0.043} & {--} & {--} & {--} \\
{\textsc{avg(}L$\geq$0)} & {0.032} & {0.034} & {--} & {0.097} & {--} & {--} & {--} \\
\bottomrule
\end{tabularx}
\end{adjustbox}
}
\vspace{-1.5mm}
\caption{Results in the CLIR task across different language pairs and word vector extraction configurations. MAP scores reported; \textit{\textsc{avg(}L$\leq$n)} means that we average representations over all Transformer layers up to the $n$th layer (included), where $L=0$ refers to the embedding layer, $L=1$ to the bottom layer, and $L=12$ to the final (top) layer. Different configurations are described in \S\ref{s:methodology} and Table~\ref{tab:config}. Additional diagnostic experiments with top-to-bottom layerwise averaging configs (\textsc{reverse}) are run for a subset of languages: \{\en, \de, \fin\}.}
\label{tab:full-clir}
\end{table*}

\begin{table*}[t]
\def\arraystretch{0.93}
\centering
{\footnotesize
\begin{adjustbox}{max width=\linewidth}
\begin{tabularx}{\textwidth}{l YY}
\toprule
{\textbf{Configuration}} & {\en} & {\de} \\
\cmidrule(lr){2-3}
\textsc{fasttext.wiki} & {$0.660_{\pm 0.008}$} & {$0.601_{\pm 0.007}$} \\
\textsc{random.xavier} & {$0.473_{\pm 0.003}$} & {$0.512_{\pm 0.008}$} \\
\cmidrule(lr){2-3}
\rowcolor{Gray}
{\monobert.\iso.\nospec} & {} & {} \\
{\textsc{avg(}L$\leq$2)} & {$0.688_{\pm 0.007}$} & {$0.649_{\pm 0.002}$} \\
{\textsc{avg(}L$\leq$4)} & {$0.698_{\pm 0.002}$} & {$0.664_{\pm 0.004}$} \\
{\textsc{avg(}L$\leq$6)} & {$0.699_{\pm 0.007}$} & {$0.677_{\pm 0.006}$} \\
{\textsc{avg(}L$\leq$8)} & {$0.706_{\pm 0.003}$} & {$0.674_{\pm 0.016}$} \\
{\textsc{avg(}L$\leq$10)} & {$0.718_{\pm 0.002}$} & {$0.679_{\pm 0.008}$} \\
{\textsc{avg(}L$\leq$12)} & {$0.714_{\pm 0.012}$} & {$0.673_{\pm 0.003}$} \\
\rowcolor{Gray}
{\monobert.\aoc-\tens.\nospec} & {} & {} \\
{\textsc{avg(}L$\leq$2)}  & {$0.690_{\pm 0.007}$} & {$0.657_{\pm 0.005}$} \\
{\textsc{avg(}L$\leq$4)} & {$0.705_{\pm 0.006}$} & {$0.671_{\pm 0.009}$} \\
{\textsc{avg(}L$\leq$6)} & {$0.714_{\pm 0.008}$} & {$0.675_{\pm 0.014}$} \\
{\textsc{avg(}L$\leq$8)} & {$0.722_{\pm 0.004}$} & {$0.681_{\pm 0.010}$} \\
{\textsc{avg(}L$\leq$10)} & {$0.719_{\pm 0.007}$} & {$0.682_{\pm 0.007}$} \\
{\textsc{avg(}L$\leq$12)} & {$0.720_{\pm 0.005}$} & {$0.680_{\pm 0.007}$} \\
\rowcolor{Gray}
{\monobert.\aoc-\huns.\nospec} & {} & {} \\
{\textsc{avg(}L$\leq$2)} & {$0.692_{\pm 0.007}$} & {$0.655_{\pm 0.007}$} \\
{\textsc{avg(}L$\leq$4)} & {$0.709_{\pm 0.007}$} & {$0.670_{\pm 0.005}$} \\
{\textsc{avg(}L$\leq$6)} & {$0.718_{\pm 0.009}$} & {$0.672_{\pm 0.008}$} \\
{\textsc{avg(}L$\leq$8)} & {$0.717_{\pm 0.003}$} & {$0.680_{\pm 0.006}$} \\
{\textsc{avg(}L$\leq$10)} & {$0.721_{\pm 0.009}$} & {$0.678_{\pm 0.004}$} \\
{\textsc{avg(}L$\leq$12)} & {$0.715_{\pm 0.003}$} & {$0.678_{\pm 0.006}$} \\
\rowcolor{Gray}
{\monobert.\iso.\all} & {} & {} \\
{\textsc{avg(}L$\leq$2)} & {$0.688_{\pm 0.008}$} & {$0.654_{\pm 0.012}$} \\
{\textsc{avg(}L$\leq$4)} & {$0.698_{\pm 0.011}$} & {$0.662_{\pm 0.008}$} \\
{\textsc{avg(}L$\leq$6)} & {$0.711_{\pm 0.005}$} & {$0.664_{\pm 0.005}$} \\
{\textsc{avg(}L$\leq$8)} & {$0.709_{\pm 0.008}$} & {$0.663_{\pm 0.015}$} \\
{\textsc{avg(}L$\leq$10)} & {$0.712_{\pm 0.006}$} & {$0.669_{\pm 0.003}$} \\
{\textsc{avg(}L$\leq$12)} & {$0.704_{\pm 0.005}$} & {$0.666_{\pm 0.013}$} \\
\rowcolor{Gray}
{\monobert.\iso.\cls} & {} & {} \\
{\textsc{avg(}L$\leq$2)} & {$0.693_{\pm 0.004}$} & {$0.649_{\pm 0.016}$} \\
{\textsc{avg(}L$\leq$4)} & {$0.699_{\pm 0.004}$} & {$0.664_{\pm 0.006}$} \\
{\textsc{avg(}L$\leq$6)} & {$0.709_{\pm 0.002}$} & {$0.671_{\pm 0.006}$} \\
{\textsc{avg(}L$\leq$8)} & {$0.710_{\pm 0.003}$} & {$0.679_{\pm 0.006}$} \\
{\textsc{avg(}L$\leq$10)} & {$0.713_{\pm 0.006}$} & {$0.670_{\pm 0.007}$} \\
{\textsc{avg(}L$\leq$12)} & {$0.705_{\pm 0.005}$} & {$0.676_{\pm 0.006}$} \\
\rowcolor{Gray}
{\mbert.\iso.\nospec} & {} & {} \\
{\textsc{avg(}L$\leq$2)} & {$0.671_{\pm 0.009}$} & {$0.628_{\pm 0.013}$} \\
{\textsc{avg(}L$\leq$4)} & {$0.669_{\pm 0.006}$} & {$0.640_{\pm 0.004}$} \\
{\textsc{avg(}L$\leq$6)} & {$0.684_{\pm 0.010}$} & {$0.637_{\pm 0.009}$} \\
{\textsc{avg(}L$\leq$8)} & {$0.680_{\pm 0.005}$} & {$0.647_{\pm 0.006}$} \\
{\textsc{avg(}L$\leq$10)} & {$0.676_{\pm 0.006}$} & {$0.629_{\pm 0.008}$} \\
{\textsc{avg(}L$\leq$12)} & {$0.681_{\pm 0.005}$} & {$0.637_{\pm 0.004}$} \\
\rowcolor{Gray}
{\mbert.\aoc-\tens.\nospec} & {} & {} \\
{\textsc{avg(}L$\leq$2)} & {$0.674_{\pm 0.005}$} & {$0.635_{\pm 0.011}$} \\
{\textsc{avg(}L$\leq$4)}& {$0.681_{\pm 0.006}$} & {$0.630_{\pm 0.007}$} \\
{\textsc{avg(}L$\leq$6)} & {$0.692_{\pm 0.008}$} & {$0.649_{\pm 0.010}$} \\
{\textsc{avg(}L$\leq$8)} & {$0.695_{\pm 0.004}$} & {$0.652_{\pm 0.011}$} \\
{\textsc{avg(}L$\leq$10)} & {$0.704_{\pm 0.005}$} & {$0.657_{\pm 0.012}$} \\
{\textsc{avg(}L$\leq$12)} & {$0.702_{\pm 0.005}$} & {$0.661_{\pm 0.008}$} \\
\rowcolor{Gray}
{\mbert.\iso.\all} & {} & {} \\
{\textsc{avg(}L$\leq$2)} & {$0.674_{\pm 0.004}$} & {$0.626_{\pm 0.014}$} \\
{\textsc{avg(}L$\leq$4)}& {$0.682_{\pm 0.009}$} & {$0.640_{\pm 0.009}$} \\
{\textsc{avg(}L$\leq$6)} & {$0.680_{\pm 0.002}$} & {$0.632_{\pm 0.007}$} \\
{\textsc{avg(}L$\leq$8)} & {$0.683_{\pm 0.003}$} & {$0.638_{\pm 0.010}$} \\
{\textsc{avg(}L$\leq$10)} & {$0.678_{\pm 0.007}$} & {$0.638_{\pm 0.015}$} \\
{\textsc{avg(}L$\leq$12)} & {$0.676_{\pm 0.013}$} & {$0.636_{\pm 0.005}$} \\
\midrule
\rowcolor{Gray}
{\monobert.\iso.\nospec (\textsc{reverse})} & {} & {} \\
{\textsc{avg(}L$\geq$12)} & {$0.683_{\pm 0.007}$} & {$0.628_{\pm 0.009}$} \\
{\textsc{avg(}L$\geq$10)} & {$0.692_{\pm 0.014}$} & {$0.628_{\pm 0.008}$} \\
{\textsc{avg(}L$\geq$8)} & {$0.688_{\pm 0.016}$} & {$0.648_{\pm 0.007}$} \\
{\textsc{avg(}L$\geq$6)} & {$0.704_{\pm 0.015}$} & {$0.658_{\pm 0.006}$} \\
{\textsc{avg(}L$\geq$4)} & {$0.704_{\pm 0.008}$} & {$0.668_{\pm 0.007}$} \\
{\textsc{avg(}L$\geq$2)} & {$0.707_{\pm 0.008}$} & {$0.667_{\pm 0.004}$} \\
{\textsc{avg(}L$\geq$0)} & {$0.714_{\pm 0.012}$} & {$0.673_{\pm 0.003}$} \\
\bottomrule
\end{tabularx}
\end{adjustbox}
}
\vspace{-1.5mm}
\caption{Results in the relation prediction task (RELP) across different word vector extraction configurations. Micro-averaged $F_1$ scores reported    , obtained as averages over 5 experimental runs for each configuration; standard deviation is also reported. \textit{\textsc{avg(}L$\leq$n)} means that we average representations over all Transformer layers up to the $n$th layer (included), where $L=0$ refers to the embedding layer, $L=1$ to the bottom layer, and $L=12$ to the final (top) layer. Different configurations are described in \S\ref{s:methodology} and Table~\ref{tab:config}. \textsc{random.xavier} are 768-dim vectors for the same vocabularies, randomly initialised via Xavier initialisation \cite{Glorot:2010aistats}.}
\label{tab:full-stm}
\end{table*}

\clearpage

\end{document}